\documentclass{article}

\PassOptionsToPackage{numbers, compress}{natbib}

\usepackage[final]{neurips_2024}
\usepackage[utf8]{inputenc} 
\usepackage[T1]{fontenc}    
\usepackage{hyperref}       
\usepackage{url}            
\usepackage{booktabs}       
\usepackage{amsfonts}       
\usepackage{amsmath}
\usepackage{mathtools}
\usepackage{nicefrac}       
\usepackage{microtype}      
\usepackage{xcolor}         
\usepackage{cleveref}
\usepackage{overpic}
\usepackage{cancel}
\usepackage{orcidlink}

\definecolor{windowsblue}{HTML}{3778bf}
\definecolor{amber}{HTML}{feb308}
\definecolor{palered}{HTML}{d9544d}
\definecolor{greyish}{HTML}{808080}
\definecolor{fadedgreen}{HTML}{7bb274}
\definecolor{dustypurple}{HTML}{825f87}
\definecolor{dustyorange}{HTML}{f0833a}

\title{The Map Equation Goes Neural:\\ Mapping Network Flows with Graph Neural Networks}

\author{%
  Christopher Bl{\"o}cker\thanks{Also at Chair of Machine Learning for Complex Networks, Center for Artificial Intelligence and Data Science, Julius-Maximilians-Universit{\"a}t W{\"u}rzburg, Germany}~~\mbox{\orcidlink{0000-0001-7881-2496}}\\
  Data Analytics Group\\
  Department of Informatics\\
  University of Zurich, Switzerland\\
  \texttt{christopher.bloecker@uzh.ch} \\
  \And
  Chester Tan~\mbox{\orcidlink{0000-0003-3823-9626}}\\
  Chair of Machine Learning for Complex Networks\\
  Center for Artificial Intelligence and Data Science\\
  Julius-Maximilians-Universit{\"a}t W{\"u}rzburg, Germany\\
  \texttt{chester.tan@uni-wuerzburg.de} \\
  \And
  Ingo Scholtes\thanks{Also at Data Analytics Group, Department of Informatics, University of Zurich, Switzerland}~~\mbox{\orcidlink{0000-0003-2253-0216}}\\
  Chair of Machine Learning for Complex Networks\\
  Center for Artificial Intelligence and Data Science\\
  Julius-Maximilians-Universit{\"a}t W{\"u}rzburg, Germany\\
  \texttt{ingo.scholtes@uni-wuerzburg.de} \\
}

\begin{document}

\maketitle

\begin{abstract}
  Community detection is an essential tool for unsupervised data exploration and revealing the organisational structure of networked systems.
  With a long history in network science, community detection typically relies on objective functions, optimised with custom-tailored search algorithms, but often without leveraging recent advances in deep learning.
  Recently, first works have started incorporating such objectives into loss functions for deep graph clustering and pooling.
  We consider the map equation, a popular information-theoretic objective function for unsupervised community detection, and express it in differentiable tensor form for optimisation through gradient descent.
  Our formulation turns the map equation compatible with any neural network architecture, enables end-to-end learning, incorporates node features, and chooses the optimal number of clusters automatically, all without requiring explicit regularisation.
  Applied to unsupervised graph clustering tasks, we achieve competitive performance against state-of-the-art deep graph clustering baselines in synthetic and real-world datasets.
\end{abstract}

\section{Introduction}
Many real-world networked systems are organised in communities: groups of nodes that are more similar to each other than to the rest.
Communities provide insights into network structure at the mesoscale, revealing sub-systems by analysing link patterns.
Motivated by different research questions, several characterisations of what constitutes ``good'' communities have been proposed~\cite{FORTUNATO201075,fortunato202220}, however, neither of them is fundamentally more correct than any other.
Moreover, no single community-detection method outperforms all others on any given network~\cite{doi:10.1126/sciadv.1602548}, motivating the ongoing efforts of research on community detection.
Typically, community-detection approaches formulate an objective function that calculates a quality score for a given partition of the network's nodes into communities.
Finding the best partition is an NP-hard search problem and often involves custom heuristic algorithms that attempt to minimise their objective function~\cite{Blondel_2008,Traag2019,algorithms-infomap}.

Graph neural networks (GNNs) have enabled applying deep learning to graph-structured data by utilising the input graph as the neural network's computational graph~\cite{the-graph-neural-network-model,pmlr-v70-gilmer17a,kipf2017semisupervised}.
Typical tasks for GNNs include node labelling, graph labelling, and link prediction, all of which involve learning meaningful representations jointly from the graph's topology, the nodes' features, and, possibly, the edges' features.
Graph labelling relies on coarse-graining the graph through identifying groups of ``similar'' nodes and aggregating their links and features, also referred to as pooling~\cite{ying2018hierarchical,bianchi2020spectral,bianchi2023expressive}, which is related to graph clustering~\cite{schaeffer2007graph}, however, these two tasks have different goals.

While GNNs excel at incorporating node and edge features with graph topology, including this information is also possible but more challenging with traditional network science approaches, typically requiring modelling or adjusting objective functions and their optimisation algorithms.
On the other hand, objective functions for community detection provide precise interpretations as to why one partition is considered better than another while deep-learning-based approaches are black boxes.
Model selection in deep learning is often done through regularisation techniques or cross-validation; in contrast, objective functions that are based on the minimum description length (MDL) principle naturally implement Occam's razor, preventing overfitting and enabling principled model selection without requiring extra regularisation or cross-validation~\cite{rissanen1978modeling,grunwald2005advances}.

Here, we combine the benefits of traditional community-detection approaches and deep learning and consider the map equation, an information-theoretic objective function for community detection~\cite{rosvall2008pnas}.
By adapting the map equation for soft cluster assignments and implementing it in differentiable tensor form, we enable end-to-end optimisation of the map equation as a loss function with gradient descent and GNNs.
In analogy to the map equation's stochastic optimisation algorithm Infomap~\cite{algorithms-infomap}, we call our approach Neuromap and evaluate it against Infomap and several recent GNN-based graph clustering methods.
Applied to synthetic and real-world networks, Neuromap demonstrates competitive performance against recent deep graph clustering baselines.

Our key contributions can be summarised as follows:
\begin{enumerate}
	\item We adapt the map equation as a differentiable loss function for end-to-end deep graph clustering and propose Neuromap, a deep-learning-based alternative to the popular Infomap algorithm for unsupervised community detection with the map equation.
    Neuromap is compatible with any neural network architecture, detects overlapping communities, leverages node features for improved performance on real-world networks, and, by following the minimum description length principle, does not require explicit regularisation.
	\item We extensively evaluate Neuromap on hundreds of synthetic and ten real datasets against recent baselines paired with various neural network architectures.
    Neuromap outperforms the baseline on the synthetic networks in most settings and is amongst the best performers in seven out of ten real datasets.
    \item By choosing a higher maximum number of clusters than previous works, we show empirically that recent baselines tend to overfit and report considerably more than the ground-truth number of communities.
    Moreover, we find that choosing a small maximum number of communities is often detrimental to graph clustering performance.
\end{enumerate}

\section{Related work}

\paragraph{\textbf{Community detection.}}
\emph{Communities}, also called \emph{clusters} or \emph{modules}, are groups of nodes that are more ``similar'' to each other than to the rest, often understood as having more links inside than between groups~\cite{FORTUNATO201075,fortunato202220}.
However, this rather general characterisation leaves precise details of what constitutes a community open.
Modularity compares the observed link densities inside communities against a randomised version of the network~\cite{doi:10.1073/pnas.0601602103}.
The stochastic block model and its variants assume a latent block structure where the probability that two nodes are connected depends only on their block memberships~\cite{HOLLAND1983109,peixoto2014efficient}.
The map equation identifies communities as regions where a random walker tends to stay for a relatively long time~\cite{rosvall2008pnas,Rosvall2009}.
Traditional clustering approaches, such as k-means, group nodes based on their proximity in space, however, here we consider identifying communities from the link patterns in networked systems.
For a detailed overview of community detection in complex networks, we refer to \cite{FORTUNATO201075,fortunato202220}.

\paragraph{Minimum description length principle.}
The minimum description length principle (MDL) is a model-selection approach that formalises Occam's razor and frames learning as a compression problem~\cite{rissanen1978modeling,grunwald2005advances}.
MDL states that the best model for data $D$ is the one that compresses the data the most.
In traditional MDL, the data's two-part description length $L\left(D\right) = \min_M L\left(M\right) + L\left(D \mid M\right)$ is the smallest achievable length over all models $M$, where $L\left(M\right)$ is the model's description length, and $L\left(D \mid M\right)$ is the data's description length, given the model.
MDL has been adopted for a wide range of applications, including regularising neural networks' weights~\cite{hinton1993keeping}, investigating deep neural networks' data-compression capabilities~\cite{NEURIPS2018_3b712de4}, analysing the characteristics of datasets~\cite{pmlr-v139-perez21a}, and community detection~\cite{rosvall2008pnas,PhysRevX.4.011047}.

\paragraph{\textbf{Deep graph clustering and pooling.}}
Graph clustering has long been a research focus in machine learning~\cite{schaeffer2007graph,yu2005soft}.
Spectral approaches cluster, for example, the eigenspace of a graph's Laplacian matrix or identify communities through graph cuts~\cite{shi2000normalized,vonLuxburg2007}.
Methods based on neural embeddings involve learning node representations with, for example, DeepWalk~\cite{10.1145/2623330.2623732} or node2vec~\cite{node2vec}, followed by applying standard clustering approaches such as k-means, assuming that similar nodes are embedded at similar locations~\cite{tandon2021community,kojaku2023network}.
Other approaches include graph autoencoders~\cite{wang2017mgae,mrabah2022rethinking}, contrastive learning~\cite{ahmadi2022deep}, and self-expressiveness~\cite{bandyopadhyay2021unsupervised}.
Recently, minimum cuts~\cite{bianchi2020spectral,bianchi2023expressive}, modularity~\cite{JMLR:v24:20-998,10.1007/978-3-319-73198-8_11}, and the Bernoulli-Poisson model~\cite{shchur2019overlapping} have been integrated with GNNs as loss functions for graph pooling and clustering.
Such GNN-based approaches can incorporate graph structure as well as node and edge features in end-to-end optimisation of the clustering objective.
Inspired by pooling in convolutional neural networks, graph pooling coarse-grains links, node features, and edge features to summarise graphs, enabling GNNs with improved performance on node and graph classification tasks \citep{ying2018hierarchical,bianchi2023expressive}.
Consequently, graph pooling has become a research focus for GNNs, emphasising the importance of graph clustering as a primary objective~\citep{ying2018hierarchical,bianchi2023expressive,JMLR:v24:20-998}.
For recent surveys of deep graph clustering, we refer to~\cite{yue2022survey,xing2022comprehensive}.

\section{Background: the map equation}
The map equation~\cite{rosvall2008pnas,Rosvall2009} is an information-theoretic objective function for unsupervised community detection that follows the MDL principle \cite{rissanen1978modeling}, and has demonstrated high performance in synthetic and real networks from across domains~\cite{Lancichinetti2009,Aldecoa2013,10.1371/journal.pone.0154404}.
The map equation formulates community detection as a compression problem and uses random walks as a proxy to model dynamic processes on networks, also called \emph{flow}.
The goal is to describe the random walk as efficiently as possible by minimising its expected per-step description length -- also called \emph{codelength} -- by partitioning the network into groups of nodes, called \emph{modules}, where the random walker tends to stay for a relatively long time.
In practice, however, the map equation does not simulate random walks; instead, the codelength is calculated analytically.

\begin{figure}[b!]
	\centering
    \includegraphics[width=\linewidth]{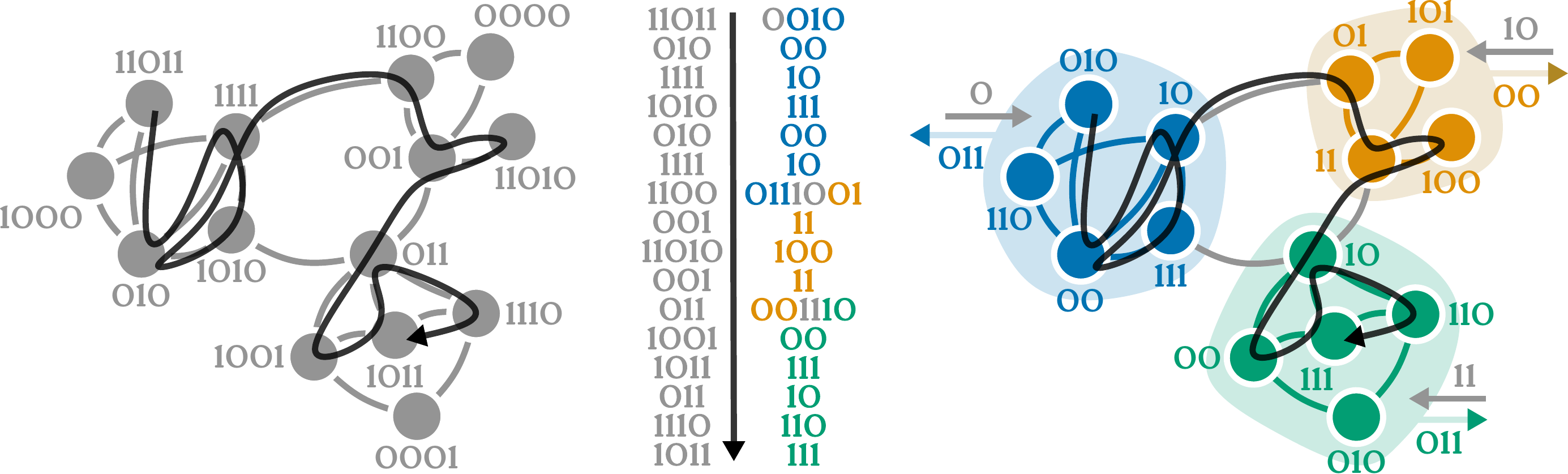}
	\caption{
		Coding principles behind the map equation.
        Colours indicate modules, codewords are shown next to nodes, and the black trace shows a sequence of random-walker steps.
		\textbf{Left:} All nodes belong to the same module and all codewords are unique.
		Encoding the random walk sequence requires $60$~bits, or $3.72$~bits per step in the limit.
		\textbf{Right:} Partitioning the network enables reusing codewords across modules, reducing the codelength.
		However, for a unique encoding, we need to introduce codewords for entering and exiting modules, shown next to the arrows pointing into and out of the modules.
		With this modular coding scheme, we can compress the description to $48$~bits, or $3.01$~bits per step in the limit.
        \textbf{Middle:} The two encodings of the random walker's steps.
	}
	\label{fig:map-equation-background}
\end{figure}

Let $G = \left(V, E\right)$ be a graph with nodes $V$, links $E$, and let $w_{uv} \in \mathbb{R}^+_0$ denote the non-negative link weight on the link from node $u$ to $v$.
When all nodes are assigned to the same module, the codelength is defined by the Shannon entropy $H$ over the nodes' visit rates~\cite{shannon1948mathematical}, $H\left(P\right) = -\sum_{u \in V} p_u \log_2 p_u$, where $p_u$ is node $u$'s visit rate and $P = \left\{p_u \,|\, u \in V\right\}$ is the set of node visit rates.
In undirected graphs, we compute visit rates directly as $p_u = s_u / \sum_{v \in V} s_v$, where $s_u = \sum_{v \in V} w_{uv}$ is node $u$'s strength.
In directed graphs, we compute the visit rates numerically with smart teleportation~\citep{lambiotte2012pre} and a power iteration.
When we partition the nodes into modules, the codelength becomes a weighted average of the modules' entropies and the entropy at the so-called index level for switching between modules.
\Cref{fig:map-equation-background} illustrates the coding principle behind the map equation using Huffman codes~\cite{huffman-coding}; note, however, that these codewords are only for illustration and we only care about their expected length in the limit to evaluate the map equation.

Minimising the map equation means balancing between small modules to achieve low module-level entropies and large modules for low index-level entropy.
This trade-off between module- and index-level entropies prevents trivial solutions where all nodes are assigned to the same module or each node is assigned to a singleton module~\cite{rissanen1978modeling}.
The map equation calculates the codelength for a partition $\mathsf{M}$,
\begin{equation}
     L\left(\mathsf{M}\right) = q H\!\left(Q\right) + \sum_{\mathsf{m} \in \mathsf{M}} p_\mathsf{m} H\!\left(P_\mathsf{m}\right).
     \label{eqn:map-equation-standard}
\end{equation}
Here, $q = \sum_\mathsf{m} q_\mathsf{m}$ is the random walker's module entry rate, $q_\mathsf{m} = \sum_{u \notin \mathsf{m}} \sum_{v \in \mathsf{m}} p_u t_{uv}$ is module $\mathsf{m}$'s entry rate, and $Q = \left\{ q_\mathsf{m} / q \mid \mathsf{m} \in \mathsf{M} \right\}$ is the set of normalised module entry rates; $p_\mathsf{m} = \mathsf{m}_\text{exit} + \sum_{u \in \mathsf{m}} p_u$ is the rate at which the random walker moves in module $\mathsf{m}$, including the module exit rate $\mathsf{m}_\text{exit} = \sum_{u \in \mathsf{m}} \sum_{v \notin \mathsf{m}} p_u t_{uv}$, and $P_\mathsf{m} = \left\{ \mathsf{m}_\text{exit} / p_\mathsf{m} \right\} \cup \left\{ p_u / p_\mathsf{m} \mid u \in \mathsf{m} \right\}$ is the set of normalised node visit and exit rates for module $\mathsf{m}$.
The random walker's transition probability from node $u$ to $v$ is $t_{uv} = w_{uv} / \sum_{v \in V} w_{uv}$.
We can rewrite the map equation as (see \Cref{appx:map-equation-rewrite})
\begin{equation}
	L\left(\mathsf{M}\right)
	= q \log_2 q
	- \!\sum_{\mathsf{m} \in \mathsf{M}} q_\mathsf{m} \log_2 q_\mathsf{m}
    - \!\sum_{\mathsf{m} \in \mathsf{M}} \mathsf{m}_\text{exit} \log_2 \mathsf{m}_\text{exit}
	  - \!\sum_{u \in V} p_u \log_2 p_u
	+ \!\sum_{\mathsf{m} \in \mathsf{M}} p_\mathsf{m} \log_2 p_\mathsf{m}. \label{eqn:map-equation}
\end{equation}

The map equation framework has been extended for overlapping communities through state-space expansions with higher-order network models~\cite{memory-networks,biased-random-walks}, avoiding over-partitioning in sparse networks using a Bayesian regularisation approach~\cite{bayesian-map-equation}, and to deal with sparse constrained structures~\cite{edler2022variable}.
Moreover, the map equation framework can incorporate node features through an extension~\cite{PhysRevE.100.022301} or by preprocessing data~\cite{doi:10.1126/sciadv.abn7558}.
Detecting communities relies on Infomap~\cite{algorithms-infomap}, a greedy stochastic search algorithm that optimises the map equation.
However, each of the above extensions requires preprocessing the input data, adjusting the loss function, or adapting the search algorithm.
In contrast, adapting the map equation as a loss function for optimisation with gradient descent does not require any custom algorithm, thus enabling flexible experimentation with variations, scalability to GPU clusters, and incorporating it into other loss functions.

\section{The map equation goes neural}
\begin{figure}[h!]
    \centering
    \includegraphics[width=\linewidth]{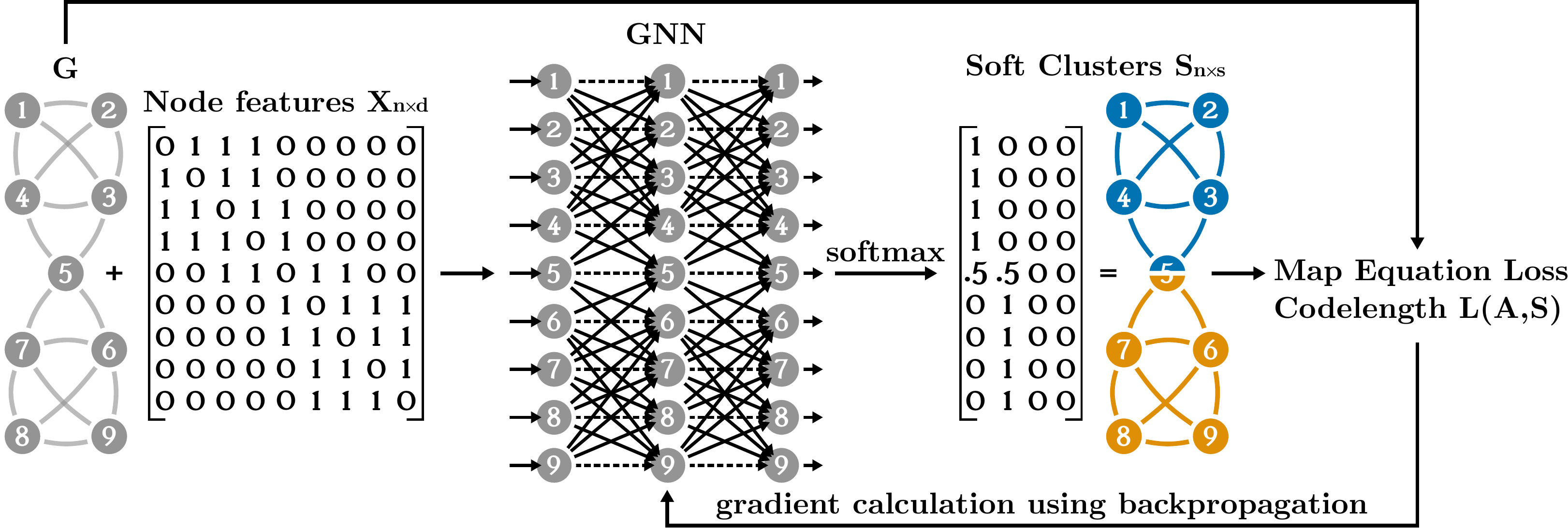}
    \caption{Illustration of the setup for GNN-based community detection with the map equation. We learn soft cluster assignments $\mathbf{S}$ from the graphs adjacency matrix $\mathbf{A}$ and the node features $\mathbf{X}$. Here, we allow up to four clusters. When no node features are available, we set $\mathbf{X} = \mathbf{A}$.}
    \label{fig:overview}
\end{figure}

We set out to detect communities by optimising the map equation with GNNs through gradient descent, which essentially means learning coarse-graining node representations in the form of communities (\Cref{fig:overview}).
While the standard map equation considers hard clusters where each node is assigned to exactly one module, we introduce a soft cluster assignment matrix $\mathbf{S}_{n \times s}$ to make the map equation differentiable and enable overlapping clusters.
We optimise $\mathbf{S} = \operatorname{softmax}\left(\operatorname{GNN}_\theta\left(\mathbf{A}, \mathbf{X}\right)\right)$ indirectly by optimising the GNN's parameters $\theta$, that is, its weights, with respect to the codelength $L$.
Here, $\mathbf{A}_{n \times n}$ is the graph's adjacency matrix, $\mathbf{X}_{n \times d}$ is the node features matrix, $n = \left|V\right|$ is the number of nodes, $s$ is the maximum allowed number of clusters, and $d$ is the node feature dimension.

Without loss of generality, we assume directed networks.
We denote the graph's total weight as $w_\text{tot} = \sum_{i \in V} \sum_{j \in V} w_{ij}$.
Let $\mathbf{T}_{n \times n}$ be the random walker's transition matrix and $\mathbf{d}^\text{in}$ be the vector of weighted node in-degrees with
\begin{equation*}
    \mathbf{T}_{ij} = \begin{dcases*}
        \frac{w_{ij}}{\sum_{j \in V} w_{ij}}
          & if $\sum_{j \in V} w_{ij} > 0$, \\
        0 & otherwise,
    \end{dcases*}
    \qquad \mathbf{d}^\text{in}_j = \sum_{i \in V} w_{ij}.
\end{equation*}
To compute the vector $\mathbf{p}$ of node visit rates, we use smart teleportation~\cite{lambiotte2012pre} and the power iteration method:
With probability $\alpha$, the random walker teleports to a random node, chosen proportionally to the nodes' in-degrees, or follows a link with probability $1-\alpha$.
This approach leads to the iterative update rule $\mathbf{p}^{(t+1)} \leftarrow \frac{\alpha}{w_\text{tot}} \mathbf{d}^\text{in} + \left(1-\alpha\right) \mathbf{p}^{(t)} \mathbf{T}$, and we set $\mathbf{p}^{(0)} = \mathbf{d}^\text{in}$.
The graph's flow matrix $\mathbf{F}_{n \times n}$ encodes the flow between each pair of nodes, where $\mathbf{F} = \frac{\alpha}{w_\text{tot}} \mathbf{A} + \left(1-\alpha\right) \operatorname{diag}\left(\mathbf{p}\right) \mathbf{T}$.
We obtain the flow $\mathbf{C}_{s \times s}$ between clusters from $\mathbf{S}$ and $\mathbf{F}$ as $\mathbf{C} = \mathbf{S}^\top \mathbf{F} \mathbf{S}$.
Following \Cref{eqn:map-equation}, we define
\begin{equation*}
    q = 1 - \operatorname{tr}\left(\mathbf{C}\right)
    \quad
    \mathbf{q}_\mathsf{m} = \mathbf{C}\mathbf{1}_s - \operatorname{diag}\left(\mathbf{C}\right)
    \quad
    \mathbf{m}_\text{exit} = (\mathbf{1}_s^\top \mathbf{C})^\top \! - \operatorname{diag}\left(\mathbf{C}\right)
    \quad
    \mathbf{p}_\mathsf{m} = \mathbf{q}_\mathsf{m} + \mathbf{1}_s^\top \mathbf{C}
\end{equation*}
and assemble the map equation
\begin{equation}
    L\left(\mathbf{A}, \mathbf{S}\right)
    = q \log_2 q
    - \hspace{-.5pt}\left(\mathbf{q}_\mathsf{m} \log_2 \mathbf{q}_\mathsf{m}\right)\mathbf{1}_s
    - \hspace{-.5pt}\left(\mathbf{m}_\text{exit} \log_2 \mathbf{m}_\text{exit}\right)\mathbf{1}_s
    - \hspace{-.5pt}\left(\mathbf{p} \log_2 \mathbf{p}\right)\mathbf{1}_n
    + \hspace{-.5pt}\left(\mathbf{p}_\mathsf{m} \log_2 \mathbf{p}_\mathsf{m}\right)\mathbf{1}_s
    \label{eqn:matrix-map-equation}
\end{equation}
where $\mathbf{1}_k$ is the $k$-dimensional vector of ones, and logarithms are applied component-wise.
The third term is constant since it only depends on the node visit rates and can be omitted during optimisation.

The map equation naturally incorporates Occam's razor by following the MDL principle for balancing between model complexity and fit~\citep{rissanen1978modeling,grunwald2005advances}, choosing the optimal number of communities automatically, but at most $s$.
In contrast, recent GNN-based clustering approaches require explicit regularisation to avoid over-partitioning \citep{ying2018hierarchical,JMLR:v24:20-998,10.1007/978-3-319-73198-8_11,shchur2019overlapping}, and our results show that they often return the maximum allowed number of communities instead of determining the number of communities in a data-driven fashion (see \Cref{sec:evaluation}).
In principle, any neural network architecture, such as a multi-layer perceptron (MLP) or GNN, can be used to learn the soft cluster assignment matrix $\mathbf{S}$.
Since the map equation involves logarithms, we add a small constant $\epsilon$ to each value in the output $\mathbf{S}$ before the backpropagation step to ensure differentiability.
We refer to the combination of using map equation loss (\Cref{eqn:matrix-map-equation}) together with a (graph) neural network to learn (overlapping) communities as \emph{Neuromap}.

\paragraph{Complexity and limitations.} The most expensive calculation is the pooling operation $\mathbf{C} = \mathbf{S}^\top \mathbf{F} \mathbf{S}$ which depends on the network's density.
When $s \ll n$ and the number of edges is $m = \mathcal{O}\left(n\right)$, the complexity of Neuromap is linear in $n$.
When the network is dense, $m = \mathcal{O}\left(n^2\right)$, or the maximum number of clusters approaches the number of nodes $ s \approx n $, we approach quadratic complexity.
Therefore, we recommend keeping $ s \ll n $ for scalability.

We assume connected networks, otherwise, clustering should be run on the individual components.
The node features $\mathbf{X}$ aid the GNN in learning patterns, however, they do not contribute to the loss.
When no node features are available, Neuromap can use, for example, the adjacency matrix as node features; designing expressive low-dimensional node features remains an active research area \citep{lim2023sign}.

\section{Experimental evaluation\label{sec:evaluation}}
We evaluate Neuromap on synthetic and real-world networks with different neural network architectures: a 2-layer graph convolutional network (GCN) \cite{kipf2017semisupervised}, a 2-layer graph isomorphism network (GIN) \cite{xu2018how}, and a 2-layer SAGE network~\cite{NIPS2017_5dd9db5e}.
To investigate whether GNNs are required for clustering, we also include a fully connected linear layer and a 2-layer MLP.
We include a learnable temperature parameter for the softmax operation, which we found speeds up convergence.
In all cases, we use the models provided by PyTorch Geometric with SELU activation~\cite{klambauer2017self}.
Because the specifics between architectures differ, such as message-passing details and aggregation functions, they may be interpreted as using different search algorithms which return different communities.
We use the Adam optimiser \cite{kingma2017adam}, apply batch normalisation, and for comparability between different methods, set the learning rate for the linear layer to $10^{-1}$, for MLP to $10^{-2}$, and for GCN, GIN, and SAGE to $10^{-3}$.
We train all models for up to $10{,}000$ epochs with a patience of $100$ epochs and dropout probability $0.5$.
Because the datasets contain hard clusters, we convert the resulting communities to hard clusters, assigning each node to that cluster where it has its strongest membership.
As baselines, we use Infomap \cite{algorithms-infomap} and five recent approaches for unsupervised graph clustering with GNNs: DMoN \cite{JMLR:v24:20-998}, NOCD \cite{shchur2019overlapping}, DiffPool \cite{ying2018hierarchical}, MinCut \cite{bianchi2020spectral}, and Ortho \cite{JMLR:v24:20-998}.
We base our implementation\footnote{\url{https://github.com/chrisbloecker/neuromap}} on PyTorch \citep{Paszke_PyTorch_An_Imperative_2019} and PyTorch Geometric \cite{fey2019fast} and ran our experiments on a workstation with an Intel i9-11900K @ 3.50GHz CPU, 32 GB of RAM, and a GeForce RTX 3090 with 24 GB of memory.

\subsection{Synthetic networks with planted communties}
We generate directed and undirected Lancichinetti-Fortunato-Radicchi (LFR) benchmark networks with planted ground-truth communities \cite{PhysRevE.78.046110} with $n = 1000$ nodes, average node degree $k \in \left\{ \ln{n}, 2\ln{n} \right\}$, maximum node degree $k_{\operatorname{max}} = 2\sqrt{n}$, both rounded to the nearest integer, and mixing parameter $\mu$ between $0.1$ and $0.8$ with a step size of $0.1$.
We set the power-law exponents for the node degree distribution to $\tau_1 = 2$, and for the community size distribution to $\tau_2 = 1$.
For each combination of parameters, we generate $10$ LFR networks using the implementation provided by the authors,\footnote{\url{https://sites.google.com/site/andrealancichinetti/benchmarks}} resulting in a total of 320 networks.
For each parameter combination, there are 10 LFR networks; for each of these LFR networks, we run each model 10 times, measuring its performance as the average adjusted mutual information (AMI)~\cite{pmlr-v32-romano14} against the ground truth, and plot the average of those AMI values over the 10 networks per parameter combination.
To verify that the number of communities is inferred from the data, we set the maximum number of communities to $s = n$.
Since LFR networks do not have node features, we use the adjacency matrix as node features.

\begin{figure*}[h!]
	\centering
    \includegraphics[width=\linewidth]{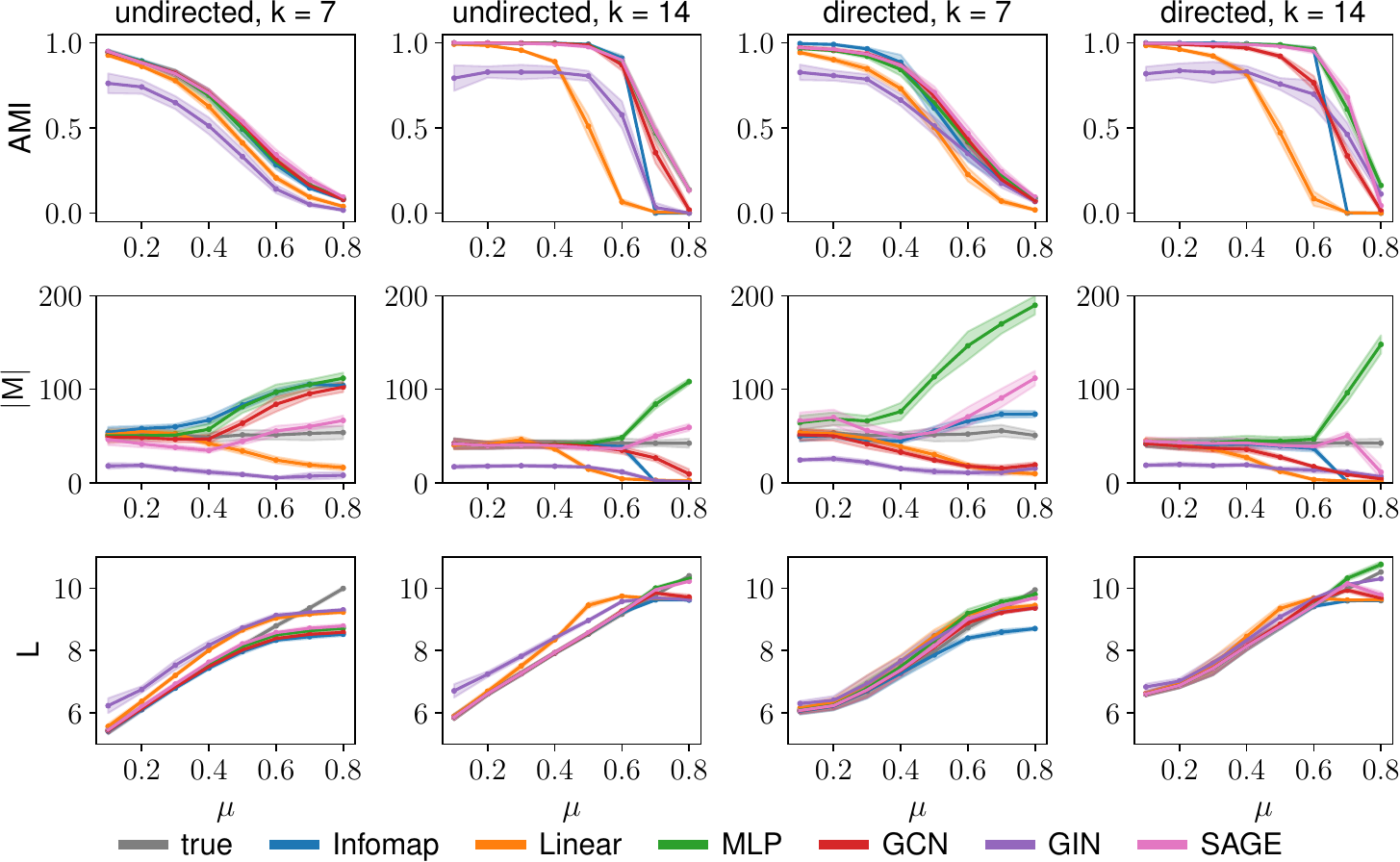}
	\caption{Performance for Neuromap using a dense linear layer, MLP, GCN, GIN, and SAGE architectures with two layers and Infomap on directed and undirected LFR networks with planted communities.
	The results show averages of partition quality measured with AMI, number of detected communities $\left|\mathsf{M}\right|$, and codelength $L$.
	The shaded areas show one standard deviation from the mean.
	}
	\label{fig:synthetic-map-equation}
\end{figure*}

We find that the detected communities' quality depends on the choice of neural network architecture (\Cref{fig:synthetic-map-equation}).
Neuromap achieves the best AMI scores with SAGE.
GCN, MLP, and Infomap perform slightly worse, however, with some variation depending on the networks' properties.
The dense linear layer and GIN show weaker performance but still identify relevant communities.
In the sparser directed networks, Infomap performs slightly better than SAGE when the mixing $\mu$ is low.
However, the AMI values do not tell the whole story:
Infomap and MLP tend to report considerably more communities than are present in the ground truth whereas the dense linear layer and GIN tend to report much fewer communities than the ground truth, especially for higher mixing values.
GCN reports more communities than are present in the ground truth in the sparser undirected networks but fewer in the directed networks.
SAGE detects close to the true number of communities in all cases.
Infomap achieves the lowest codelength across all networks.
GCN, MLP, and SAGE achieve close to Infomap's codelength, whereas the dense linear layer and GIN have slightly higher codelength.

\begin{figure*}[h!]
	\centering
    \includegraphics[width=\linewidth]{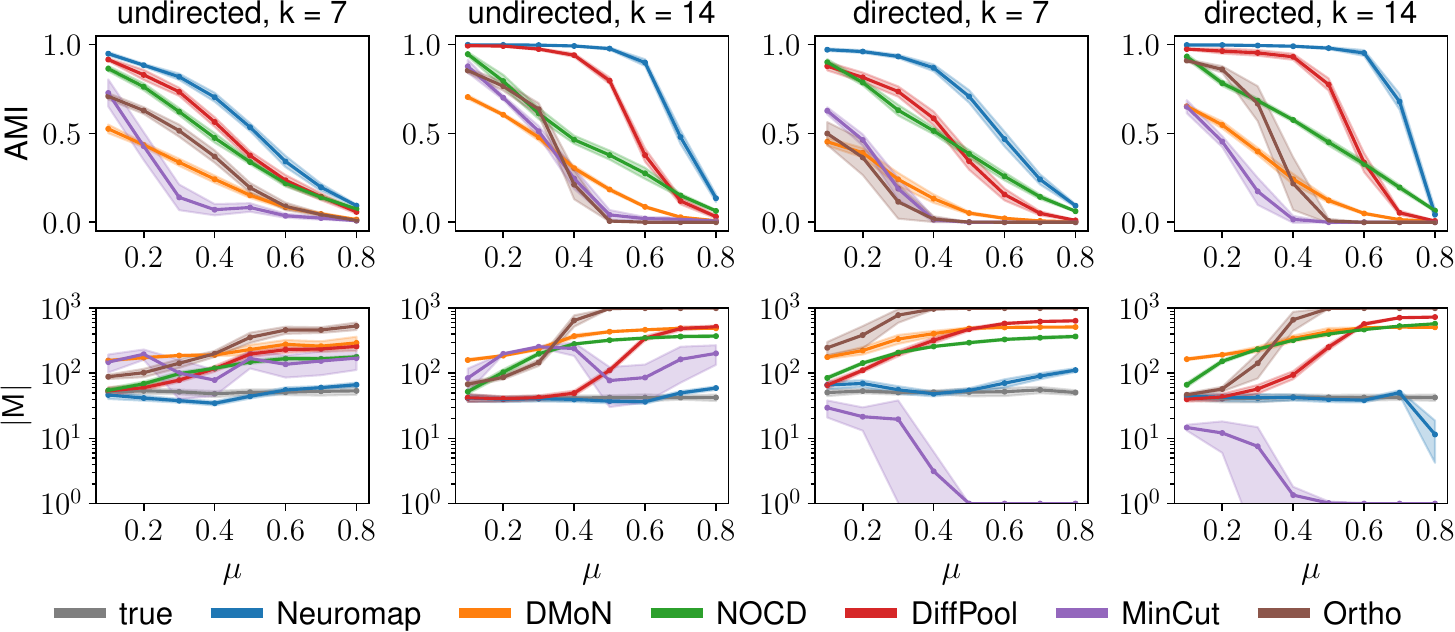}
	\caption{SAGE-based results for deep learning community-detection methods on synthetic LFR networks with planted communities.
    We show averages of partition quality measured by AMI and number of detected communities $\left|\mathsf{M}\right|$.
	The shaded areas show one standard deviation from the mean.
	}
	\label{fig:synthetic-against-baselines}
\end{figure*}

We compare Neuromap against recent deep-learning-based community detection methods on the same networks by swapping out the loss function while keeping everything else the same, with the exception of using weight decay for NOCD as per the original paper~\cite{shchur2019overlapping}.
For DiffPool, Mincut, Ortho, and DMoN, we use the implementation from PyTorch Geometric, for NOCD, we use the implementation provided by the authors.\footnote{\url{https://github.com/shchur/overlapping-community-detection}}
\Cref{fig:synthetic-against-baselines} shows the results for SAGE; in \Cref{appx:more-synthetic}, we also include results for the remaining architectures.
Neuromap outperforms the baselines across all architectures, except for NOCD which performs better than Neuromap with GIN and with GCN on directed networks.
Different from previous works, we have not limited the maximum number of communities, which allows us to analyse the methods' overfitting behaviour:
While Neuromap reports close to the ground-truth number of communities, the remaining methods often overfit the networks' structure and report considerably more communities (note the logarithmic scale).
MinCut fails to identify meaningful communities on directed networks for mixing values $\mu > 0.3$.
NOCD performs best with the GCN architecture, which was also used in the original paper \cite{shchur2019overlapping}.
Neuromap performs best with SAGE in our experiments.

\subsection{Real-world networks with node features}
\label{app:real}
We benchmark Neuromap on ten real-world datasets (\Cref{table:real-data}) from PyTorch Geometric~\cite{fey2019fast}, PyTorch Geometric Signed Directed~\cite{he2024pytorch}, and Open Graph Benchmark~\cite{hu2020open}, and compare it against the same baselines as before.
In contrast to previous works that choose a fixed number of hidden dimensions and set the maximum number of communities to a constant~\cite{JMLR:v24:20-998} or the ``ground-truth'' number of communities~\cite{shchur2019overlapping}, we reflect the networks' sizes in our choices:
We set the number of hidden dimensions to $4\sqrt{n}$ and the maximum number of communities to $s = \sqrt{n}$.
Our choices are based on empirical observations showing that the number of communities typically scales as $\mathcal{O}\left(\sqrt{n}\right)$~\cite{8692626}.
However, a few words of caution are in order: while nodes' true communities determine the link patterns in synthetic networks, it is generally infeasible to obtain ground truth communities for real networks.
Often, metadata labels are used as a drop-in, and the inferred communities' quality depends on how well the metadata, which is potentially noisy, aligns with the unknown ground truth~\cite{doi:10.1126/sciadv.1602548}.
Moreover, determining the number of communities in a network is hard and setting $s = \sqrt{n}$ should be seen as a simplification rather than an attempt to guess the exact number.

\begin{table}[t!]
	\centering
	\caption{Properties of the real-world datasets obtained from PyTorch Geometric (PyG) \cite{fey2019fast}, PyTorch Geometric Signed Directed (PyG-SD) \cite{he2024pytorch}, and Open Graph Benchmark (OGB) \cite{hu2020open}. $|V|$ is the number of nodes, $|E|$ the number of edges, $|X|$ the node feature dimension, $|Y|$ the number of communities, and $\mu$ the mixing for the given communities.}
	\begin{tabular}{lllrrrrc}
        \toprule
		Dataset              & Source & Type & $|V|$ & $|E|$ & $|X|$ & $|Y|$ & $\mu$ \\
		\midrule
		Cora                 & PyG    & Undirected &   2,708 &   5,278 & 1,433 &  7 & 0.19 \\
		CiteSeer             & PyG    & Undirected &   3,327 &   4,614 & 3,703 &  6 & 0.26 \\
		Pubmed               & PyG    & Undirected &  19,717 &  44,325 &   500 &  3 & 0.20 \\
		Amazon Computer (PC) & PyG    & Undirected &  13,752 & 143,604 &   767 & 10 & 0.22 \\
		Amazon Photo         & PyG    & Undirected &   7,650 &  71,831 &   745 &  8 & 0.17 \\
		Coauthor CS          & PyG    & Undirected &  18,333 &  81,894 & 6,805 & 15 & 0.19 \\
		Coauthor Physics     & PyG    & Undirected &  34,493 & 247,962 & 8,415 &  5 & 0.07 \\
        Cora ML              & PyG-SD & Directed   &   2,995 &   8,416 & 2,879 &  7 & 0.21 \\
        Wiki CS              & PyG-SD & Directed   &  11,701 & 297,110 &   300 & 10 & 0.31 \\
		ogb-arxiv            & OGB    & Directed   & 169,343 & 583,121 &   128 & 40 & 0.35 \\
        \bottomrule
	\end{tabular}
	\label{table:real-data}
\end{table}

\begin{figure*}[b!]
	\centering
    \includegraphics[width=\linewidth]{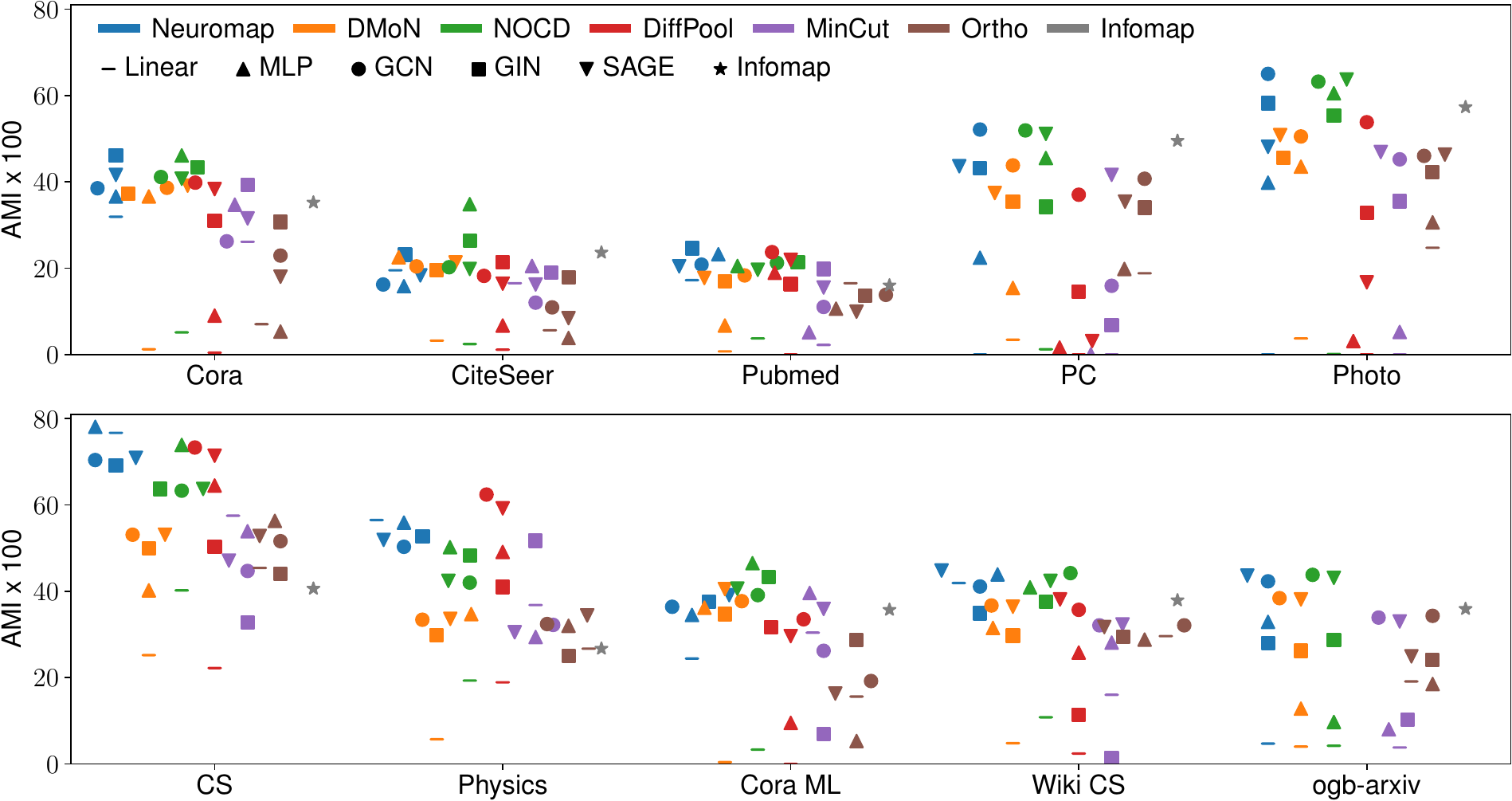}
	\caption{Average achieved AMI on real-world networks (higher is better) with $s = \sqrt{n}$. Colours indicate methods while shapes indicate neural network architectures. DiffPool ran out of memory on the ogb-arxiv dataset. Detailed tabulated results with standard deviations are included in \Cref{appx:real-world-results}.
	}
	\label{fig:real-world-results-ami}
\end{figure*}

For each method and architecture, we run 25 trials and show the average achieved AMI in \Cref{fig:real-world-results-ami}; we include a similar plot for the number of detected communities as well as the average AMI and the detected number of communities, both with standard deviations, in tabulated form in \Cref{appx:real-world-results}.
When several of the best-performing methods achieve similar AMI, we use an independent two-sample t-test to determine whether one of them can be considered to perform better than the other (see \Cref{appx:real-world-results}).
In cases where their performances do not differ significantly, we mark both as best.

Neuromap and NOCD are amongst the best performers in seven cases and DiffPool in two.
The GCN architecture performs best in seven cases, MLP and SAGE in four cases, and GIN in two cases.
A possible explanation for why the simpler linear layer and MLP architectures perform well in several cases could be that the map equation loss function captures global information in the random walker's flow patterns, making GNNs superfluous in some cases.
All methods tend to detect more communities than are present in the ``ground truth'', however, this tendency is most pronounced in DMoN, Ortho, and MinCut, which may have gone unnoticed in previous evaluations where the maximum number of communities was set to a much lower, constant value~\cite{JMLR:v24:20-998}, thus artificially preventing overfitting.
Infomap is the only baseline that does not utilise node features; instead, it relies solely on topological information, which may explain the large number of detected communities.
Comparing Neuromap's performance against Infomap's performance suggests that incorporating node features substantially improves the detected communities' quality in most cases (see \Cref{appx:real-world-results} for significance tests) while drastically reducing the number of detected communities.

We repeat the same experiments with 512 hidden features and $s = \left|Y\right|$, that is, the ``true'' number of communities, following~\cite{shchur2019overlapping} (results in \Cref{appx:fixed-arch}).
Limiting the number of allowed communities often leads to better performance for DMoN, MinCut, and Ortho, however, with a few exceptions, it diminishes the performance of Neuromap, NOCD, and DiffPool across all datasets and neural network architectures.
\Cref{appx:performance-difference} tabulates the differences in average AMI score between setting the hidden features to $4\sqrt{n}$ and $s = \sqrt{n}$ versus using $512$ hidden features and $s = \left|Y\right|$.
In the case of Neuromap, imposing a lower bound on the number of communities interferes with the MDL principle, limiting what models for the data may be explored.

\subsection{Synthetic networks with overlapping communities}
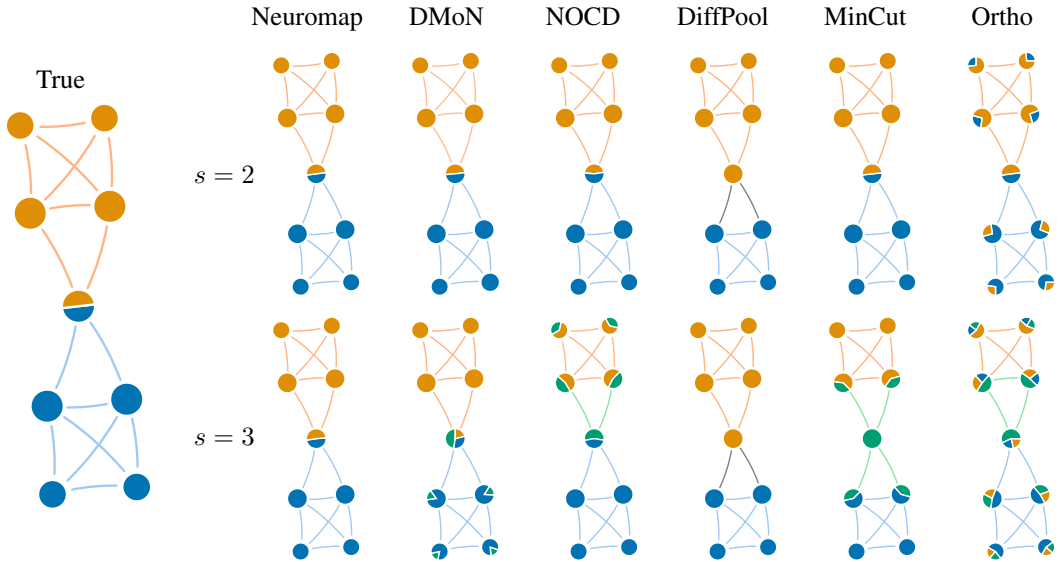
\begin{figure*}[h!]
	\centering
    \vspace*{1.5\baselineskip}
	\begin{overpic}[width=\linewidth]{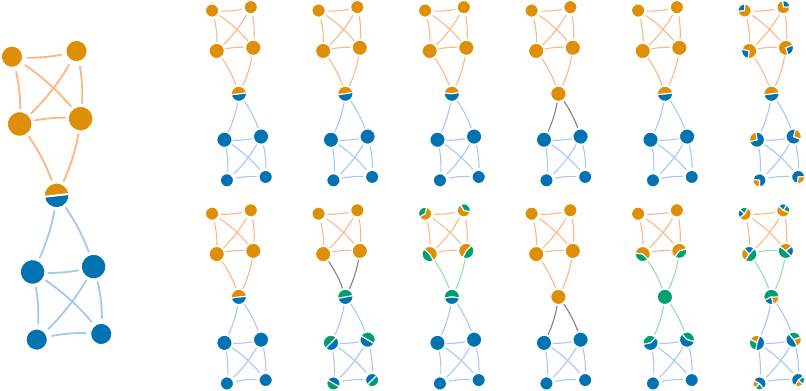}
		\put(3,45){True}
        \put(18,36){$s=2$}
        \put(18,11){$s=3$}
        \put(23.5,51){Neuromap}
        \put(38.5,51){DMoN}
        \put(51.5,51){NOCD}
        \put(64,51){DiffPool}
        \put(78,51){MinCut}
        \put(92,51){Ortho}
	\end{overpic}
	\caption{
		Synthetic network with overlapping communities where the leftmost network shows the true community structure.
        Nodes are drawn as pie charts to visualise their community assignments.
        The top and bottom rows show results for a maximum of $s = 2$ and $s = 3$ communities, respectively.
	}
	\label{fig:alcides_gcn}
\end{figure*}

We apply Neuromap and the baselines to a small synthetic network with overlapping communities~\cite{PhysRevX.1.021025}.
We set the maximum number of communities to $s \in \left\{2,3\right\}$, run each combination of loss function and neural network architecture for 10 trials, and keep the solutions with the lowest loss.
\Cref{fig:alcides_gcn} shows the results obtained with GCN, \Cref{appx:more-overlapping} shows results for the remaining architectures.

We find that Neuromap, DMoN, NOCD, and MinCut identify the correct communities for $s=2$.
DiffPool does not detect overlapping communities and Ortho assigns each node to two communities.
For $s=3$, only Neuromap identifies the correct communities.
DiffPool returns the same communities as for $s=2$.
All remaining methods return three communities.
These results provide further evidence that the baselines suffer from overfitting when they are not provided with the correct number of communities, which, in general, is unknown.
Neuromap identifies meaningful communities while inferring the number of communities in a data-driven fashion by following the MDL principle.
However, we leave a more rigorous study of overlapping communities for future work.

\section{Conclusion}
Network science and deep learning on graphs tackle community detection from different perspectives.
Community detection in network science typically relies on custom heuristic optimisation algorithms to optimise objective functions but often does not leverage recent deep learning advances.
Recently, deep graph learning methods have started to incorporate methods from network science for deep graph clustering.
We contribute to this young field by adapting the map equation, a popular unsupervised information-theoretic community-detection approach, as a differentiable loss function for end-to-end optimisation with (graph) neural networks through gradient descent, and use PyTorch to implement our approach, which we call Neuromap.

We evaluated Neuromap on various synthetic and real-world datasets, using different neural network architectures to detect communities.
Our results show that Neuromap achieves competitive performance and detects close to the ground-truth number of communities across datasets while the baselines tend to overfit and report considerably more communities.
Across all tested methods, the achieved performance depends on the used neural network architecture.
However, on several real-world benchmarks, Neuromap outperforms several of the the baselines even with simpler, non-GNN, neural network architectures.
We hypothesise that this may be because the map equation builds on capturing flow patterns, which contain global information.

While we have considered first-order networks with two-level community structures, complex real-world networks often involve higher-order dependencies and can have multi-level communities \citep{memory-networks,rosvall2011multilevelmapeq}, prompting a generalisation of our approach.
Furthermore, incorporating our method for graph pooling as well as uncovering the precise connection between the utilised neural network architecture and the achieved community-detection performance requires further empirical and theoretical studies.

\begin{ack}
We thank Martin Rosvall, Jelena Smiljani{\'c}, and Daniel Edler for fruitful discussions and helpful feedback.
We are indebted to Lisi Qarkaxhija for helping with PyTorch implementations.
Christopher Bl{\"o}cker and Ingo Scholtes acknowledge funding from the Swiss National Science Foundation, grant 176938, and the German Federal Ministry of Education and Research, grant 100582863 (TissueNet).
\end{ack}

\bibliographystyle{unsrtnat}


\clearpage
\appendix

\section{Map equation details\label{appx:map-equation-rewrite}}
Let $G = \left(V,E\right)$ be a graph with nodes $V$, links $E$, and let $w_{uv} \in \mathbb{R}^+_0$ denote the non-negative weight on the link from node $u$ to $v$.
Further, let $p_u$ be node $u$'s ergodic visit rate, which can be calculated in closed form as $p_u = s_u / \sum_{v \in V} s_v$ in undirected graphs, where $s_u = \sum_{v \in V} w_{uv}$ is node $u$'s strength.
In directed graphs, $p_u$ can be calculated numerically with smart teleportation and a power iteration by solving the recursive set of equations $p_v = \alpha \frac{s_v^\text{out}}{\sum_{u \in V} s_u^\text{out}} + \left(1-\alpha\right) \sum_{u \in V} p_u t_{uv}$, where $s_u^\text{out} = \sum_{v \in V} w_{uv}$ is node $u$'s out-strength, $t_{uv} = w_{uv} / \sum_{v \in V} w_{uv}$ is the random walker's transition probability from node $u$ to node $v$, and $\alpha$ is a teleportation parameter, typically set to $\alpha = 0.15$~\cite{lambiotte2012pre}.
Smart teleportation is similar to PageRank~\cite{gleich-pagerank-beyond-the-web} but, instead of uniformly teleporting to nodes, the random walker teleports to links proportional to their weight.

For a given partition $\mathsf{M}$ of the nodes into modules, the map equation \cite{rosvall2008pnas} calculates the average per-step description length -- also called codelength -- for describing the position of a random walker on the graph:
\begin{equation}
    L\left(\mathsf{M}\right) = q H\!\left(Q\right) + \sum_{\mathsf{m} \in \mathsf{M}} p_\mathsf{m} H\!\left(P_\mathsf{m}\right),
\end{equation}
Here, $q = \sum_\mathsf{m} q_\mathsf{m}$ is the random walker's module entry rate, $q_\mathsf{m} = \sum_{u \notin \mathsf{m}} \sum_{v \in \mathsf{m}} p_u t_{uv}$ is module $\mathsf{m}$'s entry rate, and $Q = \left\{ q_\mathsf{m} / q \mid \mathsf{m} \in \mathsf{M}\right\}$ is the set of normalised module entry rates; $p_\mathsf{m} = \mathsf{m}_\text{exit} + \sum_{u \in \mathsf{m}} p_u$ is the rate at which the random walker moves in module $\mathsf{m}$, including the exit rate $\mathsf{m}_\text{exit} = \sum_{u \in \mathsf{m}} \sum_{v \notin \mathsf{m}} p_u t_{uv}$, and $P_\mathsf{m} = \left\{ \mathsf{m}_\text{exit} / p_\mathsf{m} \right\} \cup \left\{ p_u / p_\mathsf{m} \mid u \in \mathsf{m} \right\}$ is the set of normalised node visit rates and exit rates for module $\mathsf{m}$.

\subsection{Rewriting the map equation}
The map equation can be rewritten by expanding the entropy terms and cancelling common factors
\begin{align}
    L\left(\mathsf{M}\right)
    & = q H\!\left(Q\right) + \sum_{\mathsf{m} \in \mathsf{M}} p_\mathsf{m} H\!\left(P_\mathsf{m}\right) \\
    & = - \cancel{q} \sum_{\mathsf{m} \in \mathsf{M}} \frac{q_\mathsf{m}}{\cancel{q}} \log_2 \frac{q_\mathsf{m}}{q} - \sum_{\mathsf{m} \in \mathsf{M}} \cancel{p_\mathsf{m}} \left(\frac{\mathsf{m}_\text{exit}}{\cancel{p_\mathsf{m}}} \log_2 \frac{\mathsf{m}_\text{exit}}{p_\mathsf{m}} + \sum_{u \in \mathsf{m}} \frac{p_u}{\cancel{p_\mathsf{m}}} \log_2 \frac{p_u}{p_\mathsf{m}}\right).
\end{align}
Applying logarithm rules gives
\begin{align}
    & = - \! \sum_{\mathsf{m} \in \mathsf{M}} q_\mathsf{m} \log_2 q_\mathsf{m}
        + \! \sum_{\mathsf{m} \in \mathsf{M}} q_\mathsf{m} \log_2 q \\
    & \qquad - \! \sum_{\mathsf{m} \in \mathsf{M}} \mathsf{m}_\text{exit} \log_2 \mathsf{m}_\text{exit}
        + \! \sum_{\mathsf{m} \in \mathsf{M}} \mathsf{m}_\text{exit} \log_2 p_\mathsf{m}
        - \! \sum_{\mathsf{m} \in \mathsf{M}} \sum_{u \in \mathsf{m}} p_u \log_2 p_u
        + \! \sum_{\mathsf{m} \in \mathsf{M}} \sum_{u \in \mathsf{m}} p_u \log_2 p_\mathsf{m},
\end{align}
and further simplification yields
\begin{equation}
    = q \log_2 q - \! \sum_{\mathsf{m} \in \mathsf{M}} q_\mathsf{m} \log_2 q_\mathsf{m} + \! \sum_{\mathsf{m} \in \mathsf{M}} p_\mathsf{m} \log_2 p_\mathsf{m} - \! \sum_{\mathsf{m} \in \mathsf{M}} \mathsf{m}_\text{exit} \log_2 \mathsf{m}_\text{exit} - \! \sum_{u \in V} p_u \log_2 p_u.
\end{equation}
In undirected networks where $q_\mathsf{m} = \mathsf{m}_\text{exit}$, this can be further simplified~\cite{Rosvall2009} to
\begin{equation}
    = q \log_2 q - 2 \sum_{\mathsf{m} \in \mathsf{M}} q_\mathsf{m} \log_2 q_\mathsf{m} + \! \sum_{\mathsf{m} \in \mathsf{M}} p_\mathsf{m} \log_2 p_\mathsf{m} - \! \sum_{u \in V} p_u \log_2 p_u.
    \label{eqn:map-equation-rewrite-appendix}
\end{equation}
The last term in \Cref{eqn:map-equation-rewrite-appendix}, that is, the nodes' contribution to the codelength, is constant because it does not depend on the module structure, and can be omitted during optimisation.

\subsection{Node flow with soft cluster assignments}
\begin{figure*}[t!]
	\centering
 	\begin{overpic}[width=.9\linewidth]{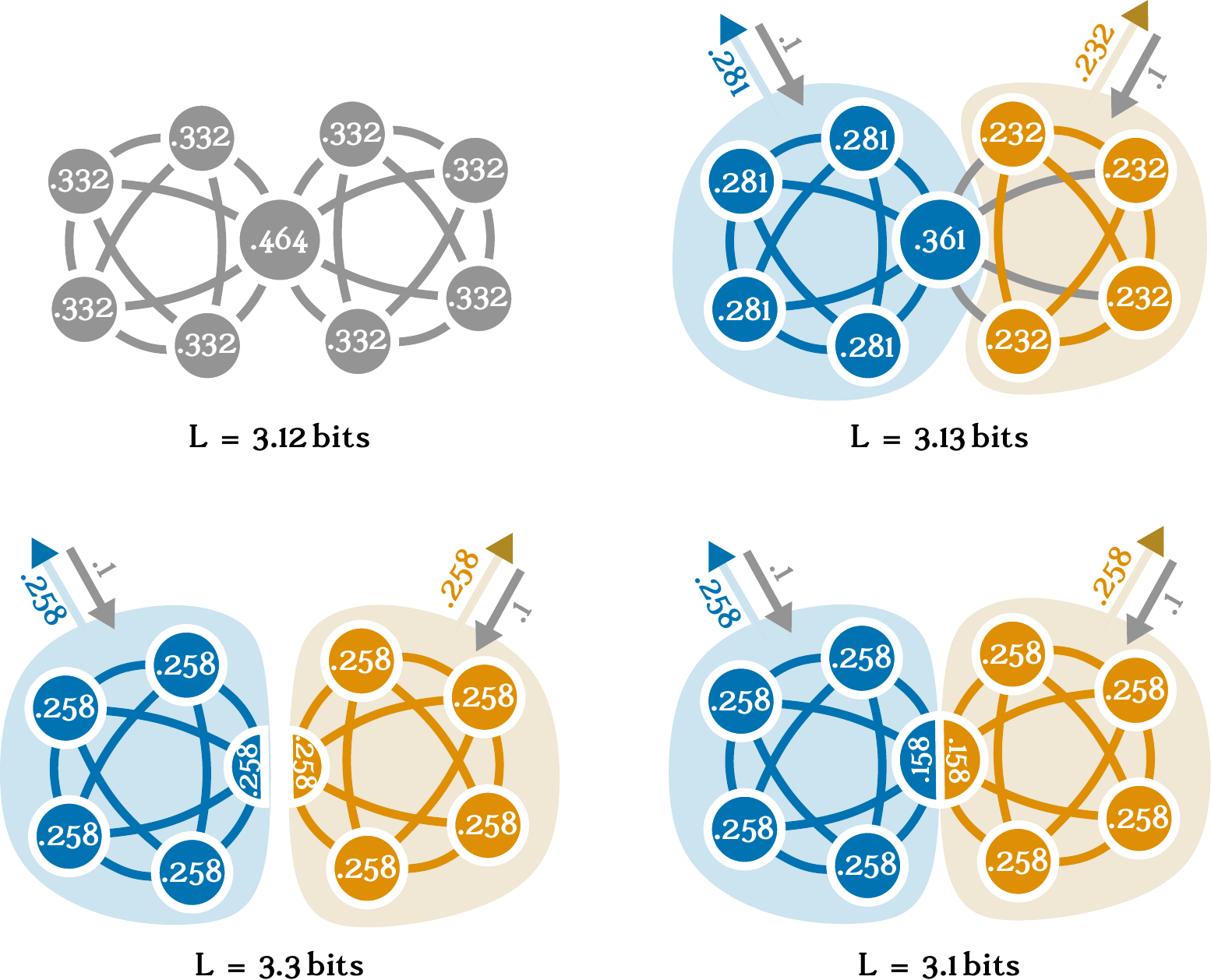}
		\put(0,75){\textbf{(a)}}
        \put(54,80){\textbf{(b)}}
        \put(-1,37){\textbf{(c)}}
        \put(54,37){\textbf{(d)}}
	\end{overpic}
	\caption{\textbf{Modelling node flow with soft cluster assignments.} The labels show each node's contribution to the overall codelength based on their visit rates.
    \textbf{(a)} The one-level partition where all nodes belong to the same community.
    \textbf{(b)} With hard communities, the middle node must be assigned to either the blue or orange community. Here, this leads to a higher codelength than for the one-level partition.
    \textbf{(c)} Reflecting nodes' partial assignments in the codelength contributions means splitting them into several smaller pieces whose visit rates sum to the original node's visit rate. Important objects are split into several less important objects, requiring longer codewords because of their lower visit rates. Overall, this tends to increase the codelength and can prevent identifying overlapping modules.
    \textbf{(d)} Treating nodes' contribution to the codelength as constant keeps the nodes intact and allows important nodes to retain their higher visit rates, leading to shorter codewords. As the codelength highlights, only this approach would identify communities in this example.
	}
	\label{fig:overlapping-node-flow}
\end{figure*}

While the nodes' codelength contribution is constant with hard clusters (\Cref{eqn:map-equation-rewrite-appendix}), expressing the map equation with soft cluster assignments allows modelling the flow of nodes that are assigned to more than one module in at least two different ways which we illustrate in \Cref{fig:overlapping-node-flow}.

The first option is to reflect the nodes' partial cluster assignment for assigning codewords.
Consider node $u$ with visit rate $p_u$ and partial assignments of $\frac{1}{2}$ to clusters $\mathsf{m}_1$ whose usage rate is $p_{\mathsf{m}_1}$ and $\frac{1}{2}$ to cluster $\mathsf{m}_2$ whose usage rate is $p_{\mathsf{m}_2}$.
Then, half of $u$'s flow, that is $\frac{p_u}{2}$, falls into each of the two clusters and $u$'s contribution to the overall codelength is $- \frac{p_u}{2} \log_2 \frac{p_u}{2 p_{\mathsf{m}_1}} - \frac{p_u}{2} \log_2 \frac{p_u}{p_{2 \mathsf{m}_2}}$.
In general, let node $u$'s assignment to cluster $\mathsf{m}_i$ be $s_{ui} \in \left[0,1\right]$ with $\sum_i s_{ui} = 1$, and let $\mathsf{m}_i$'s usage rate be $p_{\mathsf{m}_i}$.
Then, node $u$'s contribution to the overall codelength is $- \sum_i s_{ui} p_u \log_2 \frac{s_{ui} p_u}{p_{\mathsf{m}_i}}$.
Essentially, this approach splits each node with assignments to more than one cluster into its corresponding parts, resulting in potentially many small pieces with low visit rates.

The second option is to treat the nodes as indivisible when computing their codeword lengths but reflecting their soft assignments in the contribution to the overall codelength.
Using the same example as above, node $u$'s contribution to the overall codelength is $- \frac{p_u}{2} \log_2 \frac{p_u}{p_{\mathsf{m}_1}} - \frac{p_u}{2} \log_2 \frac{p_u}{p_{\mathsf{m}_2}}$, which is lower because the values inside the logarithms are bigger.
In general, $u$'s contribution to the codelength is $- \sum_i s_{ui} p_u \log_2 \frac{p_u}{p_{\mathsf{m}_i}}$, where the difference to the previous option is that $s_{ui}$ does not appear inside the logarithm.

\clearpage
\section{Results on synthetic networks with different (G)NN architectures\label{appx:more-synthetic}}
Here, we report further results on the synthetic LFR networks with planted communities for Neuromap, DMoN, NOCD, DiffPool, MinCut, and Ortho using the following architectures: a dense linear layer, a 2-layer MLP, a 2-layer GCN, and a 2-layer GIN.
The setup is as described in the main text.

\subsection{Linear-layer-based results}
\begin{figure*}[h!]
	\centering
    \includegraphics[width=\linewidth]{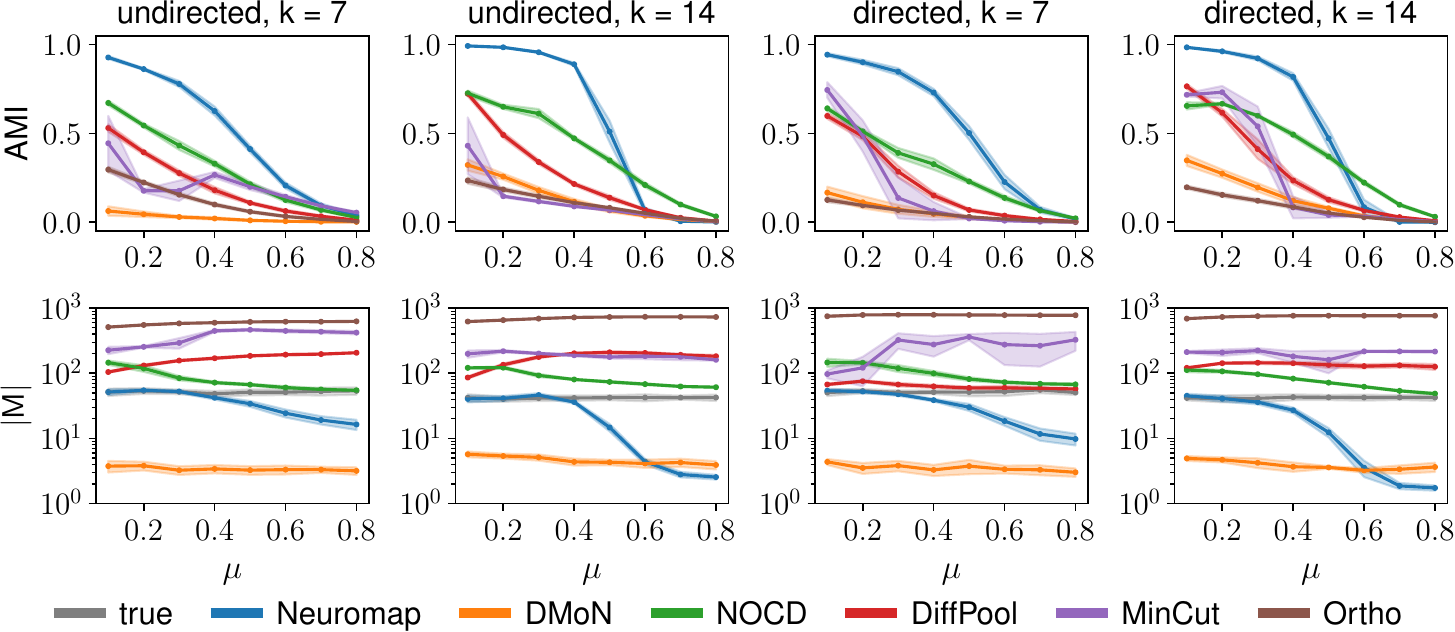}
	\caption{Linear-layer-based results for deep-learning community-detection methods on synthetic LFR networks with planted communities.
    The results show averages of partition quality measured by AMI and number of detected communities $\left|\mathsf{M}\right|$.
	The shaded areas show one standard deviation from the mean.
	}
	\label{fig:synthetic-against-baselines-linear}
\end{figure*}

\subsection{MLP-based results}
\begin{figure*}[h!]
	\centering
    \includegraphics[width=\linewidth]{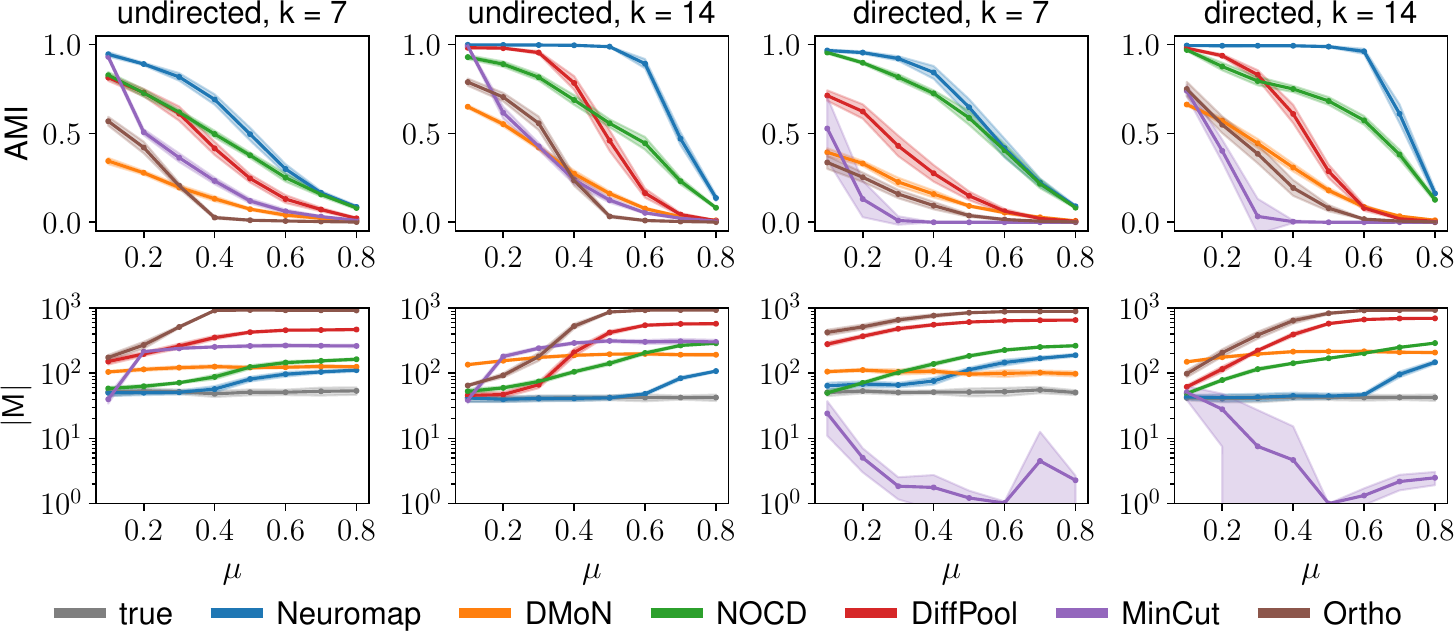}
	\caption{MLP-based results for deep-learning community-detection methods on synthetic LFR networks with planted communities.
    The results show averages of partition quality measured by AMI and number of detected communities $\left|\mathsf{M}\right|$.
	The shaded areas show one standard deviation from the mean.
	}
	\label{fig:synthetic-against-baselines-mlp}
\end{figure*}

\clearpage

\subsection{GCN-based results}
\begin{figure*}[h!]
	\centering
    \includegraphics[width=\linewidth]{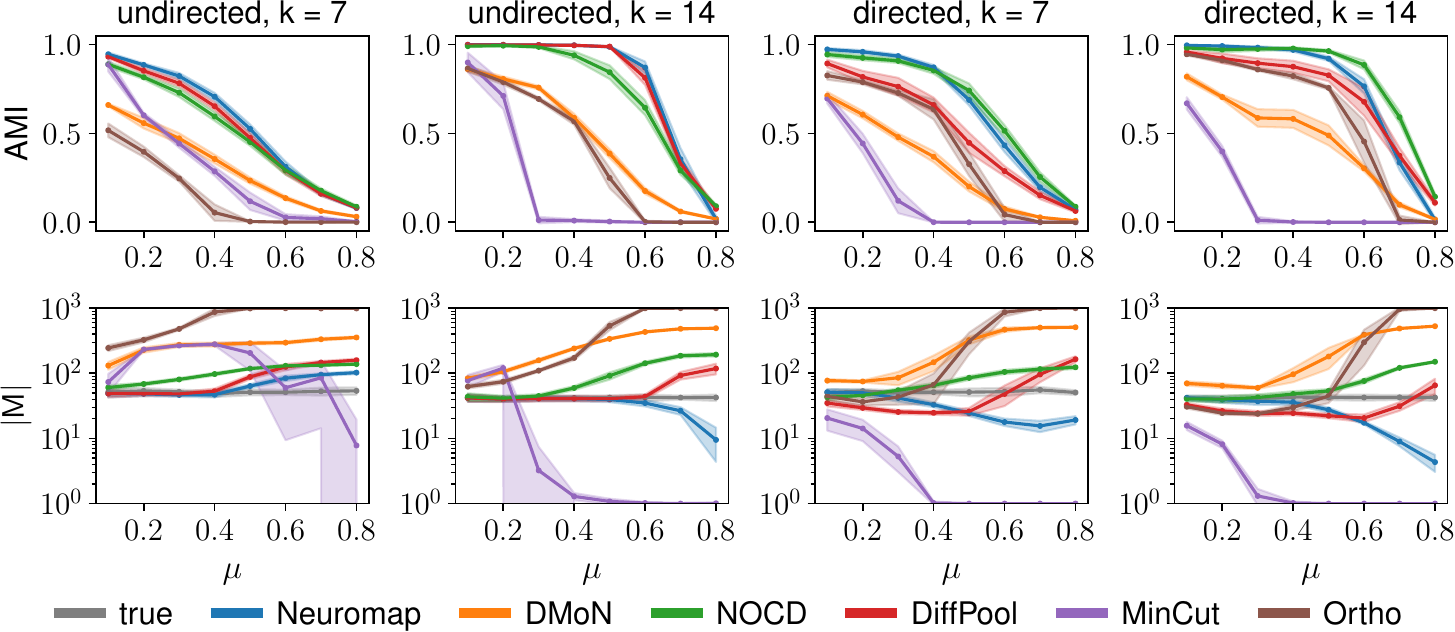}
	\caption{GCN-based results for deep-learning community-detection methods on synthetic LFR networks with planted communities.
    The results show averages of partition quality measured by AMI and number of detected communities $\left|\mathsf{M}\right|$.
	The shaded areas show one standard deviation from the mean.
	}
	\label{fig:synthetic-against-baselines-gcn}
\end{figure*}

\subsection{GIN-based results}
\begin{figure*}[h!]
	\centering
    \includegraphics[width=\linewidth]{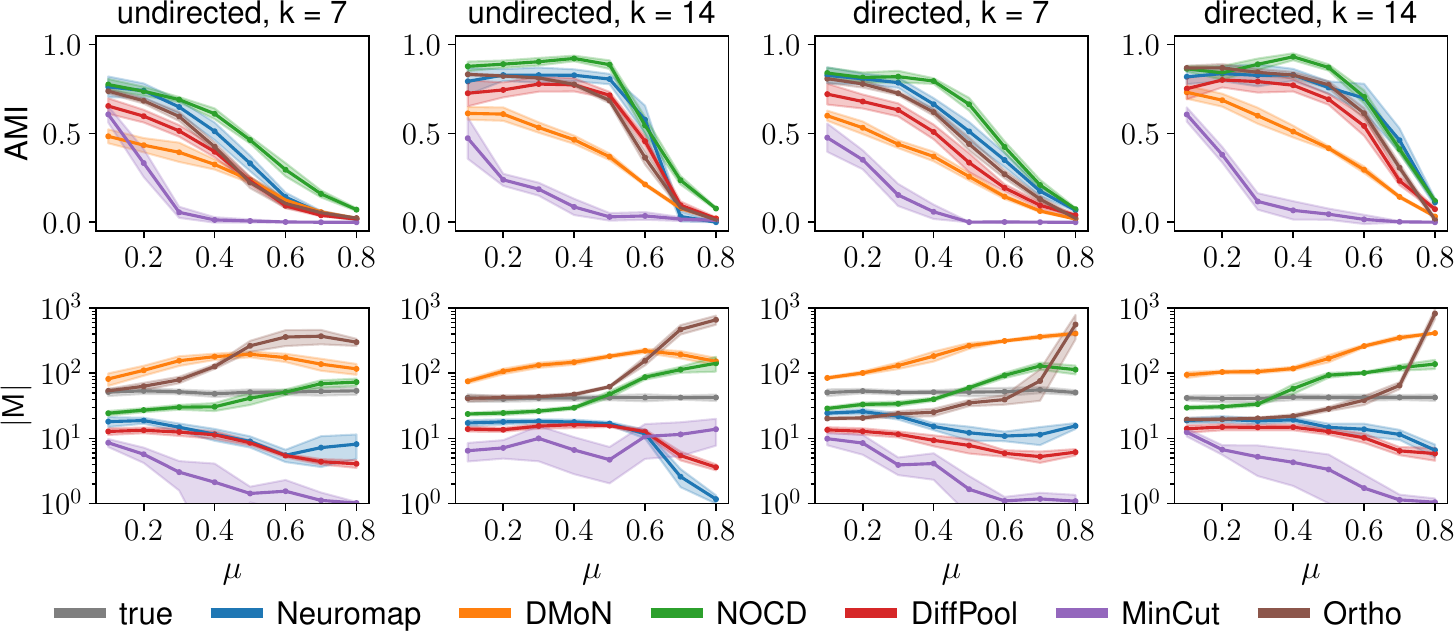}
	\caption{GIN-based results for deep-learning community-detection methods on synthetic LFR networks with planted communities.
    The results show averages of partition quality measured by AMI and number of detected communities $\left|\mathsf{M}\right|$.
	The shaded areas show one standard deviation from the mean.
	}
	\label{fig:synthetic-against-baselines-gin}
\end{figure*}

\clearpage
\section{Results on Real-World Networks\label{appx:real-world-results}}
Here, we report further results on real-world networks for setting the number of hidden dimensions to $4\sqrt{n}$ and the maximum number of communities to $s = \sqrt{n}$.
Specifically, we visualise the average detected number of communities per loss function and neural network architecture for each dataset.
We tabulate the average AMI and average number of detected communities, including standard deviations.
Two-sample t-tests show when Neuromap or a baseline performs significantly better than the other.

\subsection{Number of detected communities on real-world networks}
\begin{figure*}[h!]
	\centering
    \includegraphics[width=\linewidth]{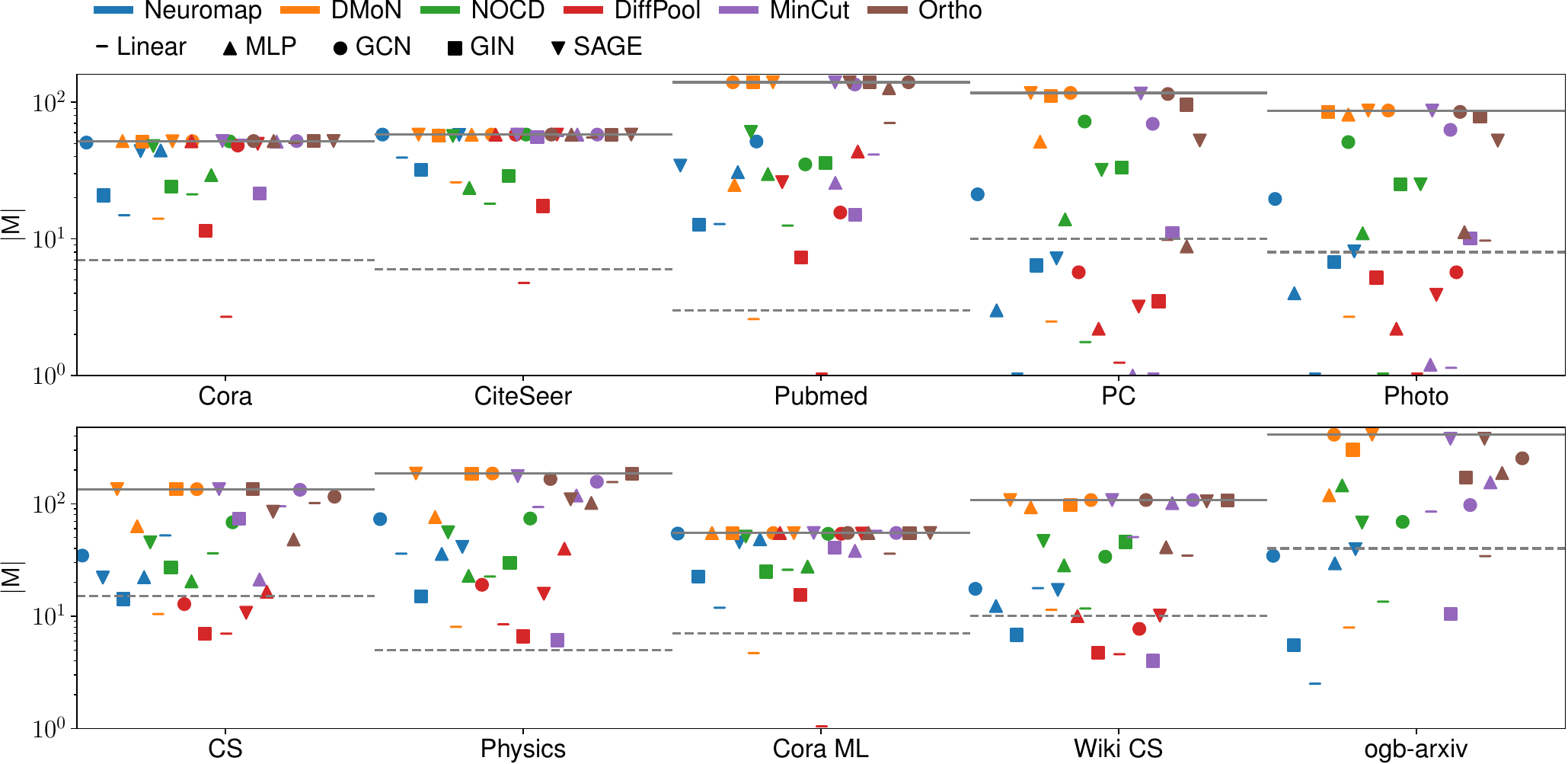}
	\caption{Average detected number of communities on real-world networks. Colours indicate methods while shapes indicate (G)NN architectures. The dashed horizontal lines show the correct number of communities for each dataset while the solid horizontal lines show the chosen maximum allowed number of communities, here $s = \sqrt{n}$. DiffPool ran out of memory on the ogb-arxiv dataset. We omit Infomap in the plot due to the large number of detected communities.
	}
	\label{fig:real-world-results-m}
\end{figure*}

\clearpage
\subsection{Tabulated AMI results on real-world networks}

Here, we report the detailed AMI performance for Neuromap, DMoN, NOCD, DiffPool, MinCut, Ortho, and Infomap on real-world networks for setting the number of hidden dimensions to $4\sqrt{n}$ and the maximum number of communities to $s = \sqrt{n}$.
When several of the best-performing methods achieve similar AMI, we use an independent two-sample t-test to determine whether one of them can be considered to perform better than the other.
In cases where their performances do not differ significantly, we mark both as best.

\newcommand{\tb}[1]{\mathbf{#1}}
\newcommand{\tu}[1]{\underline{#1}}
\newcommand{\ti}[1]{\textit{#1}}

\newcommand{\fatblue}[1]{\textcolor{windowsblue}{\mathbf{#1}}}
\newcommand{\fatred}[1]{\textcolor{palered}{\mathbf{#1}}}

\begin{table*}[h!]
	\centering
	\caption{Average AMI in \% (higher is better) and their standard deviations on real-world networks.
	For each dataset, we highlight the best scores in bold and underline the second-best score.
    We tested each method with 2-layer MLP, GCN, GIN, and SAGE architectures, except for Infomap, which is not based on deep learning.
    OOM stands for ``out of memory''.
	}
    \resizebox{\linewidth}{!}{%
	\begin{tabular}{llcccccccccc}
        \toprule
		Method & Arch. & Cora & CiteSeer & PubMed & PC & Photo & CS & Physics & Cora ML & Wiki CS & ogb-arxiv \\
        \midrule
        Neuromap & LIN & $31.9 \pm 7.9$ & $19.5 \pm 1.6$ & $17.2 \pm 12.2$ & $0.0 \pm 0.0$ & $0.0 \pm 0.0$ & $\tu{76.7 \pm 1.4}$ & $56.5 \pm 1.2$ & $24.4 \pm 11.9$ & $41.9 \pm 2.3$ & $4.7 \pm 7.7$ \\
         & MLP & $36.6 \pm 2.4$ & $15.8 \pm 1.1$ & $23.2 \pm 1.8$ & $22.4 \pm 8.6$ & $39.8 \pm 11.2$ & $\tb{78.1 \pm 2.0}$ & $55.9 \pm 2.2$ & $34.5 \pm 1.6$ & $43.9 \pm 2.4$ & $32.9 \pm 1.3$ \\
         & GCN & $38.5 \pm 2.0$ & $16.2 \pm 1.3$ & $20.8 \pm 1.2$ & $\tb{52.1 \pm 2.3}$ & $\tb{65.0 \pm 3.8}$ & $70.4 \pm 1.6$ & $50.3 \pm 1.4$ & $36.4 \pm 1.2$ & $41.1 \pm 3.2$ & $42.3 \pm 1.7$ \\
         & GIN & $\tb{46.1 \pm 2.0}$ & $23.1 \pm 1.2$ & $\tb{24.6 \pm 1.7}$ & $43.2 \pm 2.4$ & $58.2 \pm 6.0$ & $69.1 \pm 2.8$ & $52.7 \pm 2.7$ & $37.5 \pm 1.9$ & $34.9 \pm 4.5$ & $28.0 \pm 2.8$ \\
         & SAGE & $41.6 \pm 1.9$ & $18.3 \pm 1.2$ & $20.4 \pm 1.0$ & $43.6 \pm 4.8$ & $48.1 \pm 5.8$ & $70.9 \pm 2.4$ & $51.9 \pm 2.2$ & $39.1 \pm 2.1$ & $\tb{44.8 \pm 2.2}$ & $\tb{43.6 \pm 1.0}$ \\
        \midrule
        DMoN & LIN & $1.2 \pm 1.6$ & $3.2 \pm 1.6$ & $0.7 \pm 1.0$ & $3.4 \pm 2.9$ & $3.7 \pm 4.0$ & $25.2 \pm 10.8$ & $5.7 \pm 7.3$ & $0.4 \pm 0.8$ & $4.8 \pm 3.9$ & $4.0 \pm 3.6$ \\
         & MLP & $36.6 \pm 1.2$ & $22.5 \pm 0.9$ & $6.7 \pm 4.1$ & $15.4 \pm 11.6$ & $43.5 \pm 8.9$ & $40.2 \pm 6.5$ & $34.7 \pm 3.5$ & $36.2 \pm 1.2$ & $31.5 \pm 8.0$ & $12.8 \pm 1.7$ \\
         & GCN & $38.6 \pm 0.9$ & $20.4 \pm 0.9$ & $18.3 \pm 0.2$ & $43.8 \pm 0.3$ & $50.5 \pm 0.3$ & $53.1 \pm 0.2$ & $33.4 \pm 0.2$ & $37.7 \pm 0.9$ & $36.7 \pm 0.4$ & $38.4 \pm 0.2$ \\
         & GIN & $37.2 \pm 1.1$ & $19.5 \pm 0.9$ & $17.0 \pm 0.3$ & $35.4 \pm 2.3$ & $45.6 \pm 0.8$ & $49.9 \pm 0.4$ & $29.8 \pm 0.4$ & $34.7 \pm 1.5$ & $29.7 \pm 0.8$ & $26.2 \pm 3.2$ \\
         & SAGE & $39.1 \pm 1.0$ & $21.3 \pm 1.3$ & $17.7 \pm 0.2$ & $37.4 \pm 11.9$ & $50.8 \pm 0.3$ & $53.1 \pm 0.3$ & $33.6 \pm 0.2$ & $40.5 \pm 1.0$ & $36.4 \pm 0.4$ & $38.1 \pm 0.3$ \\
        \midrule
        NOCD & LIN & $5.1 \pm 1.9$ & $2.4 \pm 1.3$ & $3.7 \pm 1.7$ & $1.2 \pm 2.7$ & $0.1 \pm 0.3$ & $40.2 \pm 9.8$ & $19.3 \pm 7.0$ & $3.3 \pm 1.1$ & $10.8 \pm 3.1$ & $4.2 \pm 2.5$ \\
         & MLP & $\tb{46.1 \pm 1.1}$ & $\tb{34.8 \pm 0.9}$ & $20.5 \pm 2.4$ & $45.5 \pm 2.5$ & $60.5 \pm 3.4$ & $73.9 \pm 1.7$ & $50.2 \pm 1.0$ & $\tb{46.5 \pm 1.4}$ & $40.9 \pm 1.2$ & $9.7 \pm 2.3$ \\
         & GCN & $41.1 \pm 0.9$ & $20.2 \pm 0.8$ & $21.2 \pm 0.6$ & $\tb{51.9 \pm 1.9}$ & $63.2 \pm 0.9$ & $63.3 \pm 1.0$ & $42.0 \pm 0.7$ & $39.1 \pm 1.0$ & $\tb{44.2 \pm 0.6}$ & $\tb{43.8 \pm 0.4}$ \\
         & GIN & $43.4 \pm 1.1$ & $\tu{26.4 \pm 1.2}$ & $21.3 \pm 0.6$ & $34.2 \pm 1.9$ & $55.3 \pm 1.9$ & $63.7 \pm 0.9$ & $48.3 \pm 1.5$ & $43.3 \pm 1.2$ & $37.6 \pm 1.1$ & $28.7 \pm 1.2$ \\
         & SAGE & $40.7 \pm 0.9$ & $19.8 \pm 0.8$ & $19.6 \pm 0.4$ & $\tb{51.1 \pm 1.6}$ & $\tb{63.7 \pm 1.6}$ & $63.7 \pm 0.8$ & $42.4 \pm 0.7$ & $40.6 \pm 1.1$ & $42.4 \pm 0.4$ & $43.1 \pm 0.3$ \\
        \midrule
        DiffPool & LIN & $0.4 \pm 1.1$ & $1.1 \pm 1.9$ & $0.0 \pm 0.0$ & $0.0 \pm 0.0$ & $0.0 \pm 0.0$ & $22.2 \pm 14.3$ & $18.9 \pm 15.9$ & $0.0 \pm 0.1$ & $2.4 \pm 2.1$ & OOM \\
         & MLP & $9.0 \pm 1.2$ & $6.7 \pm 1.0$ & $18.9 \pm 1.4$ & $1.6 \pm 0.4$ & $3.1 \pm 0.9$ & $64.5 \pm 4.8$ & $49.1 \pm 4.4$ & $9.5 \pm 1.2$ & $25.8 \pm 2.6$ & OOM \\
         & GCN & $39.8 \pm 2.0$ & $18.2 \pm 1.3$ & $\tb{23.7 \pm 1.5}$ & $37.0 \pm 8.0$ & $53.8 \pm 7.5$ & $73.3 \pm 2.0$ & $\tb{62.4 \pm 3.6}$ & $33.5 \pm 1.2$ & $35.7 \pm 3.8$ & OOM \\
         & GIN & $30.9 \pm 3.6$ & $21.3 \pm 2.3$ & $16.3 \pm 3.9$ & $14.5 \pm 3.2$ & $32.8 \pm 4.6$ & $50.3 \pm 3.9$ & $41.0 \pm 3.9$ & $31.6 \pm 2.8$ & $11.3 \pm 3.1$ & OOM \\
         & SAGE & $38.3 \pm 2.4$ & $16.4 \pm 1.4$ & $21.9 \pm 1.5$ & $3.1 \pm 0.3$ & $16.7 \pm 6.1$ & $71.4 \pm 3.4$ & $\tu{59.2 \pm 3.3}$ & $29.6 \pm 2.2$ & $38.1 \pm 2.8$ & OOM \\
        \midrule
        MinCut & LIN & $26.1 \pm 2.5$ & $16.5 \pm 1.3$ & $2.2 \pm 1.0$ & $0.0 \pm 0.0$ & $0.0 \pm 0.0$ & $57.5 \pm 0.6$ & $36.8 \pm 0.9$ & $30.4 \pm 1.2$ & $16.0 \pm 1.4$ & $3.8 \pm 3.5$ \\
         & MLP & $34.7 \pm 6.6$ & $20.5 \pm 0.9$ & $5.1 \pm 2.4$ & $0.0 \pm 0.0$ & $5.2 \pm 9.9$ & $53.9 \pm 6.4$ & $29.4 \pm 9.9$ & $39.6 \pm 1.3$ & $28.1 \pm 4.2$ & $8.0 \pm 2.7$ \\
         & GCN & $26.2 \pm 1.5$ & $12.0 \pm 0.7$ & $11.0 \pm 2.8$ & $15.9 \pm 21.2$ & $45.2 \pm 13.6$ & $44.7 \pm 3.3$ & $32.2 \pm 6.8$ & $26.2 \pm 0.7$ & $32.1 \pm 0.5$ & $33.9 \pm 4.7$ \\
         & GIN & $39.3 \pm 3.8$ & $18.9 \pm 0.7$ & $19.8 \pm 4.3$ & $6.8 \pm 6.8$ & $35.5 \pm 19.3$ & $32.7 \pm 19.5$ & $51.7 \pm 5.4$ & $6.9 \pm 4.5$ & $1.3 \pm 6.2$ & $10.3 \pm 7.3$ \\
         & SAGE & $31.5 \pm 1.5$ & $16.2 \pm 0.8$ & $15.5 \pm 0.3$ & $41.6 \pm 0.6$ & $46.9 \pm 0.5$ & $47.1 \pm 0.5$ & $30.5 \pm 2.4$ & $35.9 \pm 0.7$ & $32.3 \pm 0.5$ & $33.0 \pm 1.5$ \\
        \midrule
        Ortho & LIN & $7.0 \pm 0.7$ & $5.6 \pm 0.7$ & $16.5 \pm 0.4$ & $18.8 \pm 9.0$ & $24.7 \pm 11.4$ & $45.4 \pm 2.0$ & $26.7 \pm 1.5$ & $15.6 \pm 1.1$ & $29.6 \pm 0.9$ & $19.1 \pm 0.5$ \\
         & MLP & $5.3 \pm 0.5$ & $3.8 \pm 0.4$ & $10.6 \pm 0.4$ & $19.8 \pm 5.5$ & $30.6 \pm 4.0$ & $56.3 \pm 1.7$ & $32.0 \pm 1.1$ & $5.3 \pm 0.5$ & $28.8 \pm 0.9$ & $18.5 \pm 0.4$ \\
         & GCN & $22.9 \pm 0.9$ & $10.9 \pm 0.7$ & $13.8 \pm 0.5$ & $40.7 \pm 0.9$ & $46.0 \pm 1.4$ & $51.6 \pm 1.0$ & $32.4 \pm 1.0$ & $19.2 \pm 1.1$ & $32.1 \pm 0.5$ & $34.3 \pm 0.7$ \\
         & GIN & $30.7 \pm 1.3$ & $17.8 \pm 0.9$ & $13.6 \pm 0.5$ & $34.0 \pm 0.9$ & $42.3 \pm 0.7$ & $44.1 \pm 0.6$ & $25.0 \pm 0.7$ & $28.7 \pm 1.1$ & $29.4 \pm 0.6$ & $24.1 \pm 0.8$ \\
         & SAGE & $18.0 \pm 0.9$ & $8.4 \pm 0.8$ & $9.9 \pm 0.4$ & $35.4 \pm 3.9$ & $46.3 \pm 1.6$ & $52.8 \pm 1.2$ & $34.4 \pm 1.1$ & $16.3 \pm 1.1$ & $31.7 \pm 0.5$ & $24.9 \pm 0.3$ \\
        \midrule
        Infomap &  & $35.2 \pm 0.2$ & $23.6 \pm 0.1$ & $16.0 \pm 0.1$ & $49.5 \pm 0.5$ & $57.3 \pm 1.1$ & $40.6 \pm 0.2$ & $26.7 \pm 0.1$ & $35.7 \pm 0.3$ & $37.9 \pm 0.2$ & $35.9 \pm 0.1$ \\
        \bottomrule
	\end{tabular}
    }
	\label{table:real-ami}
\end{table*}

\subsection{Significance of best Neuromap results vs. best baseline}
\begin{table}[h!]
	\centering
	\caption{Independent two-sample t-test between 25 samples of AMI values for Neuromap vs. the best baseline for $4\sqrt{n}$ hidden features and $s = \sqrt{n}$. The p-values indicate for which datasets the null hypothesis ``the samples have the same mean'' can be rejected. \textcolor{windowsblue}{Blue p-values} highlight cases where Neuromap performs significantly better than the best baseline; \textcolor{palered}{red p-values} highlight cases where the best baseline performs significantly better than Neuromap.}
	\begin{tabular}{llclcc}
        \toprule
		Dataset   & \multicolumn{2}{c}{Best Neuromap} & \multicolumn{2}{c}{Best Baseline} & p \\
		\midrule
		Cora & GIN & $46.1 \pm 2.0$ & NOCD MLP & $46.1 \pm 1.1$ & $0.92$ \\
		CiteSeer & GIN & $23.1 \pm 1.2$ & NOCD MLP & $34.8 \pm 0.9$ & $\fatred{3.3e^{-35}}$ \\
		PubMed & GIN & $24.6 \pm 1.7$ & DiffPool GCN & $23.7 \pm 1.5$ & $0.05$ \\
		PC & GCN & $52.1 \pm 2.3$ & NOCD GCN & $51.9 \pm 1.9$ & $0.69$ \\
		Photo & GCN & $65.0 \pm 3.8$ & NOCD SAGE & $63.7 \pm 1.6$ & $0.14$ \\
		CS & MLP & $78.1 \pm 2.0$ & NOCD MLP & $73.9 \pm 1.7$ & $\fatblue{4.4e^{-10}}$ \\
		Physics & LIN & $56.5 \pm 1.2$ & DiffPool GCN & $62.4 \pm 3.6$ & $\fatred{1.5e^{-08}}$ \\
        Cora ML & SAGE & $39.1 \pm 2.1$ & NOCD MLP & $46.5 \pm 1.4$ & $\fatred{5.7e^{-18}}$ \\
        Wiki CS & SAGE & $44.8 \pm 2.2$ & NOCD GCN & $44.2 \pm 0.6$ & $0.23$ \\
		ogb-arxiv & SAGE & $43.6 \pm 1.0$ & NOCD GCN & $43.8 \pm 0.4$ & $0.38$ \\
        \bottomrule
	\end{tabular}
	\label{table:significance-best-neuromap-vs-baselines}
\end{table}

\clearpage
\subsection{Significance of Neuromap results vs. Infomap}
\begin{table*}[h!]
	\centering
	\caption{Independent two-sample t-test between 25 samples of AMI values for Neuromap with different (G)NN architectures vs. Infomap for $4\sqrt{n}$ hidden features and $s = \sqrt{n}$. The p-values indicate when the null hypothesis ``the samples have the same mean'' can be rejected. \textcolor{windowsblue}{Blue p-values} highlight cases where Neuromap performs significantly better than Infomap; \textcolor{palered}{red p-values} highlight cases where Infomap performs significantly better than Neuromap.
	}
    \begin{tabular}{lccccc}
        \toprule
		Dataset & LIN & MLP & GCN & GIN & SAGE \\
        \midrule
        Cora      & $0.05$ & $\fatblue{6.8e^{-3}}$ & $\fatblue{2.4e^{-8}}$ & $\fatblue{8.1e^{-20}}$ & $\fatblue{9.4e^{-15}}$ \\
        CiteSeer  & $\fatred{7.5e^{-12}}$ & $\fatred{4.2e^{-22}}$ & $\fatred{2.8e^{-20}}$ & $0.06$ & $\fatred{2.1e^{-17}}$ \\
        PubMed    & $0.65$ & $\fatblue{2.1e^{-16}}$ & $\fatblue{4.2e^{-16}}$ & $\fatblue{2.5e^{-18}}$ & $\fatblue{2.2e^{-17}}$ \\
        PC        & $\fatred{1.8e^{-48}}$ & $\fatred{4.7e^{-14}}$ & $\fatblue{6.0e^{-6}}$ & $\fatred{2.4e^{-14}}$ & $\fatred{3.9e^{-6}}$ \\
        Photo     & $\fatred{1.8e^{-42}}$ & $\fatred{6.6e^{-8}}$ & $\fatblue{2.3e^{-10}}$ & $0.48$ & $\fatred{5.2e^{-8}}$ \\
        CS        & $\fatblue{7.6e^{-36}}$ & $\fatblue{1.1e^{-32}}$ & $\fatblue{2.9e^{-32}}$ & $\fatblue{7.5e^{-26}}$ & $\fatblue{5.1e^{-28}}$ \\
        Physics   & $\fatblue{2.7e^{-35}}$ & $\fatblue{1.2e^{-28}}$ & $\fatblue{5.8e^{-31}}$ & $\fatblue{3.2e^{-25}}$ & $\fatblue{2.7e^{-27}}$ \\
        Cora ML   & $\fatred{9.6e^{-5}}$ & $\fatred{1.4e^{-3}}$ & $0.01$ & $\fatblue{7.0e^{-5}}$ & $\fatblue{1.7e^{-8}}$ \\
        Wiki CS   & $\fatblue{7.0e^{-9}}$ & $\fatblue{8.3e^{-12}}$ & $\fatblue{4.5e^{-5}}$ & $\fatred{3.9e^{-3}}$ & $\fatblue{3.4e^{-14}}$ \\
        ogb-arxiv & $\fatred{1.8e^{-16}}$ & $\fatred{1.3e^{-11}}$ & $\fatblue{5.4e^{-16}}$ & $\fatred{4.6e^{-13}}$ & $\fatblue{8.9e^{-23}}$ \\
        \bottomrule
	\end{tabular}
	\label{table:significance-neuromap-vs-infomap}
\end{table*}

\subsection{Tabulated number of detected communities on real-world networks}\label{appx:real-m}
Here, we report the average number of detected communities for Neuromap, DMoN, NOCD, DiffPool, MinCut, Ortho, and Infomap on real-world networks for setting the number of hidden dimensions to $4\sqrt{n}$ and the maximum number of communities to $s = \sqrt{n}$.

\begin{table*}[h!]
	\centering
	\caption{Average number of detected communities and their standard deviations on real-world networks.
    We tested each method with 2-layer MLP, GCN, GIN, and SAGE architectures, except for Infomap, which is not based on deep learning.
    OOM stands for ``out of memory''.
	}
    \resizebox{\linewidth}{!}{%
	\begin{tabular}{llcccccccccc}
        \toprule
		Method & Arch. & Cora & CiteSeer & PubMed & PC & Photo & CS & Physics & Cora ML & Wiki CS & ogb-arxiv \\
        \midrule
        Neuromap & LIN & $14.4 \pm 3.7$ & $38.0 \pm 3.9$ & $12.4 \pm 8.1$ & $1.0 \pm 0.0$ & $1.0 \pm 0.0$ & $50.1 \pm 5.2$ & $34.5 \pm 4.5$ & $11.4 \pm 2.5$ & $17.0 \pm 1.7$ & $2.4 \pm 1.6$ \\
         & MLP & $44.4 \pm 2.5$ & $57.3 \pm 0.9$ & $30.8 \pm 3.8$ & $3.0 \pm 0.8$ & $4.0 \pm 1.2$ & $22.2 \pm 2.4$ & $35.8 \pm 3.3$ & $48.3 \pm 1.9$ & $12.3 \pm 1.3$ & $29.5 \pm 4.4$ \\
         & GCN & $50.7 \pm 0.9$ & $58.0 \pm 0.0$ & $51.6 \pm 4.4$ & $21.2 \pm 3.4$ & $19.6 \pm 2.1$ & $34.6 \pm 4.5$ & $73.0 \pm 8.0$ & $54.3 \pm 0.7$ & $17.5 \pm 2.4$ & $34.4 \pm 7.0$ \\
         & GIN & $20.8 \pm 2.2$ & $32.0 \pm 2.7$ & $12.7 \pm 1.3$ & $6.4 \pm 1.3$ & $6.8 \pm 0.8$ & $14.2 \pm 2.6$ & $14.9 \pm 1.8$ & $22.6 \pm 2.7$ & $6.8 \pm 1.7$ & $5.5 \pm 1.3$ \\
         & SAGE & $43.9 \pm 2.4$ & $57.6 \pm 0.6$ & $34.4 \pm 2.5$ & $7.2 \pm 1.9$ & $8.1 \pm 1.3$ & $22.0 \pm 2.8$ & $41.3 \pm 6.1$ & $45.4 \pm 2.8$ & $17.1 \pm 2.0$ & $39.5 \pm 4.4$ \\
        \midrule
        DMoN & LIN & $13.6 \pm 13.3$ & $25.1 \pm 10.8$ & $2.5 \pm 1.7$ & $2.4 \pm 1.0$ & $2.6 \pm 1.4$ & $10.0 \pm 6.3$ & $7.7 \pm 6.7$ & $4.5 \pm 6.2$ & $10.9 \pm 5.6$ & $7.6 \pm 6.4$ \\
         & MLP & $52.0 \pm 0.2$ & $57.9 \pm 0.3$ & $24.8 \pm 33.0$ & $51.4 \pm 36.3$ & $81.2 \pm 15.0$ & $63.1 \pm 17.1$ & $76.3 \pm 6.7$ & $55.0 \pm 0.2$ & $92.9 \pm 27.1$ & $118.9 \pm 13.2$ \\
         & GCN & $51.9 \pm 0.3$ & $58.0 \pm 0.0$ & $140.0 \pm 0.0$ & $117.0 \pm 0.0$ & $87.0 \pm 0.0$ & $135.0 \pm 0.0$ & $186.0 \pm 0.0$ & $54.8 \pm 0.4$ & $108.0 \pm 0.0$ & $412.0 \pm 0.0$ \\
         & GIN & $51.6 \pm 0.6$ & $57.0 \pm 1.0$ & $140.0 \pm 0.0$ & $110.6 \pm 6.1$ & $84.5 \pm 1.7$ & $134.9 \pm 0.3$ & $185.9 \pm 0.3$ & $54.5 \pm 0.6$ & $97.2 \pm 4.0$ & $301.8 \pm 77.3$ \\
         & SAGE & $51.8 \pm 0.4$ & $58.0 \pm 0.0$ & $140.0 \pm 0.2$ & $116.8 \pm 0.7$ & $86.8 \pm 0.4$ & $135.0 \pm 0.0$ & $186.0 \pm 0.0$ & $54.8 \pm 0.4$ & $108.0 \pm 0.0$ & $412.0 \pm 0.0$ \\
        \midrule
        NOCD & LIN & $20.5 \pm 10.2$ & $17.5 \pm 21.5$ & $12.1 \pm 3.1$ & $1.7 \pm 1.0$ & $1.0 \pm 0.2$ & $34.7 \pm 8.5$ & $21.6 \pm 2.4$ & $24.8 \pm 3.9$ & $11.2 \pm 0.9$ & $12.9 \pm 2.8$ \\
         & MLP & $29.3 \pm 2.7$ & $23.6 \pm 4.1$ & $29.8 \pm 2.7$ & $13.9 \pm 3.9$ & $11.0 \pm 2.5$ & $20.4 \pm 3.7$ & $22.8 \pm 2.7$ & $27.6 \pm 1.7$ & $28.3 \pm 2.0$ & $145.4 \pm 41.7$ \\
         & GCN & $51.7 \pm 0.6$ & $58.0 \pm 0.0$ & $35.1 \pm 2.9$ & $72.3 \pm 19.2$ & $51.1 \pm 11.0$ & $68.4 \pm 8.6$ & $73.9 \pm 7.3$ & $54.0 \pm 0.8$ & $33.8 \pm 3.9$ & $69.0 \pm 6.5$ \\
         & GIN & $24.2 \pm 1.7$ & $28.8 \pm 2.6$ & $35.9 \pm 6.2$ & $33.1 \pm 5.8$ & $25.1 \pm 4.0$ & $27.0 \pm 4.1$ & $29.9 \pm 4.3$ & $24.9 \pm 2.6$ & $45.7 \pm 5.1$ & $78.5 \pm 15.6$ \\
         & SAGE & $47.8 \pm 1.6$ & $56.6 \pm 1.3$ & $60.6 \pm 3.3$ & $32.0 \pm 3.0$ & $25.1 \pm 3.0$ & $45.3 \pm 3.6$ & $55.7 \pm 2.9$ & $51.1 \pm 1.7$ & $46.6 \pm 2.7$ & $68.0 \pm 6.6$ \\
        \midrule
        DiffPool & LIN & $2.6 \pm 1.1$ & $4.6 \pm 3.3$ & $1.0 \pm 0.0$ & $1.2 \pm 0.5$ & $1.0 \pm 0.0$ & $6.7 \pm 6.1$ & $8.1 \pm 5.5$ & $1.0 \pm 0.2$ & $4.4 \pm 1.0$ & OOM \\
         & MLP & $51.8 \pm 0.4$ & $58.0 \pm 0.2$ & $43.6 \pm 4.7$ & $2.2 \pm 0.5$ & $2.2 \pm 0.4$ & $16.5 \pm 2.6$ & $39.8 \pm 6.1$ & $55.0 \pm 0.0$ & $10.0 \pm 1.4$ & OOM \\
         & GCN & $48.2 \pm 1.6$ & $57.9 \pm 0.4$ & $15.6 \pm 2.1$ & $5.7 \pm 1.1$ & $5.7 \pm 0.8$ & $12.8 \pm 2.3$ & $19.0 \pm 3.3$ & $54.1 \pm 0.9$ & $7.7 \pm 1.4$ & OOM \\
         & GIN & $11.5 \pm 2.2$ & $17.4 \pm 3.1$ & $7.3 \pm 1.6$ & $3.5 \pm 0.6$ & $5.2 \pm 0.8$ & $7.0 \pm 1.1$ & $6.6 \pm 1.1$ & $15.5 \pm 2.5$ & $4.7 \pm 1.0$ & OOM \\
         & SAGE & $49.7 \pm 1.3$ & $58.0 \pm 0.2$ & $26.0 \pm 3.0$ & $3.2 \pm 0.4$ & $3.9 \pm 0.7$ & $10.7 \pm 1.6$ & $15.8 \pm 2.1$ & $54.2 \pm 0.7$ & $10.1 \pm 1.2$ & OOM \\
        \midrule
        MinCut & LIN & $51.3 \pm 0.9$ & $54.8 \pm 1.9$ & $40.0 \pm 6.1$ & $1.0 \pm 0.2$ & $1.1 \pm 0.3$ & $91.0 \pm 3.9$ & $89.6 \pm 6.2$ & $55.0 \pm 0.0$ & $48.5 \pm 4.9$ & $81.6 \pm 93.0$ \\
         & MLP & $51.5 \pm 1.6$ & $58.0 \pm 0.0$ & $25.7 \pm 7.4$ & $1.0 \pm 0.0$ & $1.2 \pm 0.4$ & $21.2 \pm 41.8$ & $118.6 \pm 24.6$ & $38.1 \pm 5.8$ & $101.4 \pm 19.2$ & $155.0 \pm 49.5$ \\
         & GCN & $52.0 \pm 0.0$ & $58.0 \pm 0.0$ & $134.9 \pm 4.1$ & $69.5 \pm 25.2$ & $62.8 \pm 20.3$ & $133.2 \pm 2.9$ & $157.0 \pm 40.2$ & $55.0 \pm 0.0$ & $108.0 \pm 0.0$ & $97.4 \pm 20.9$ \\
         & GIN & $21.5 \pm 20.0$ & $55.7 \pm 8.5$ & $15.0 \pm 37.5$ & $11.0 \pm 4.2$ & $10.1 \pm 5.5$ & $73.9 \pm 63.8$ & $6.1 \pm 6.7$ & $40.6 \pm 6.0$ & $4.0 \pm 14.5$ & $10.5 \pm 4.2$ \\
         & SAGE & $52.0 \pm 0.0$ & $58.0 \pm 0.0$ & $140.0 \pm 0.0$ & $115.9 \pm 2.8$ & $86.9 \pm 0.3$ & $135.0 \pm 0.0$ & $176.5 \pm 24.6$ & $55.0 \pm 0.0$ & $108.0 \pm 0.0$ & $379.5 \pm 33.7$ \\
        \midrule
        Ortho & LIN & $48.7 \pm 1.3$ & $53.4 \pm 1.7$ & $68.1 \pm 5.7$ & $9.5 \pm 4.1$ & $9.4 \pm 5.1$ & $97.4 \pm 6.6$ & $149.5 \pm 5.4$ & $34.5 \pm 3.3$ & $33.2 \pm 2.6$ & $32.8 \pm 3.4$ \\
         & MLP & $52.0 \pm 0.0$ & $58.0 \pm 0.0$ & $126.9 \pm 4.7$ & $8.8 \pm 2.3$ & $11.2 \pm 2.0$ & $48.1 \pm 9.1$ & $101.7 \pm 11.2$ & $55.0 \pm 0.0$ & $41.0 \pm 6.2$ & $187.7 \pm 18.9$ \\
         & GCN & $52.0 \pm 0.0$ & $58.0 \pm 0.0$ & $140.0 \pm 0.0$ & $114.9 \pm 1.6$ & $85.0 \pm 3.3$ & $115.6 \pm 11.9$ & $166.2 \pm 15.9$ & $55.0 \pm 0.0$ & $108.0 \pm 0.0$ & $253.8 \pm 68.3$ \\
         & GIN & $52.0 \pm 0.0$ & $58.0 \pm 0.0$ & $140.0 \pm 0.0$ & $95.8 \pm 4.4$ & $78.8 \pm 2.8$ & $135.0 \pm 0.0$ & $185.8 \pm 1.0$ & $55.0 \pm 0.0$ & $107.5 \pm 0.8$ & $170.2 \pm 22.4$ \\
         & SAGE & $52.0 \pm 0.0$ & $58.0 \pm 0.0$ & $140.0 \pm 0.0$ & $52.5 \pm 30.2$ & $52.4 \pm 21.7$ & $84.9 \pm 17.5$ & $109.5 \pm 18.1$ & $55.0 \pm 0.0$ & $105.1 \pm 1.6$ & $379.8 \pm 4.6$ \\
        \midrule
        Infomap &  & $282.2 \pm 4.1$ & $628.9 \pm 2.7$ & $924.9 \pm 9.1$ & $455.1 \pm 5.1$ & $223.5 \pm 4.4$ & $834.4 \pm 9.4$ & $1160.8 \pm 11.4$ & $287.4 \pm 3.8$ & $747.0 \pm 8.4$ & $4840.0 \pm 22.4$ \\
        \bottomrule
	\end{tabular}
    }
	\label{table:real-m}
\end{table*}

\clearpage
\section{Results on real-world networks with fixed number of hidden channels\label{appx:fixed-arch}}
We provide further results on real-world networks with a different configuration of the neural networks: we set the number of hidden channels to 512 and set the maximum number of communities to the ``ground-truth'' number of communities, that is $s = \left|Y\right|$.

\subsection{Average Achieved AMI}
\begin{figure*}[h!]
	\centering
    \includegraphics[width=\linewidth]{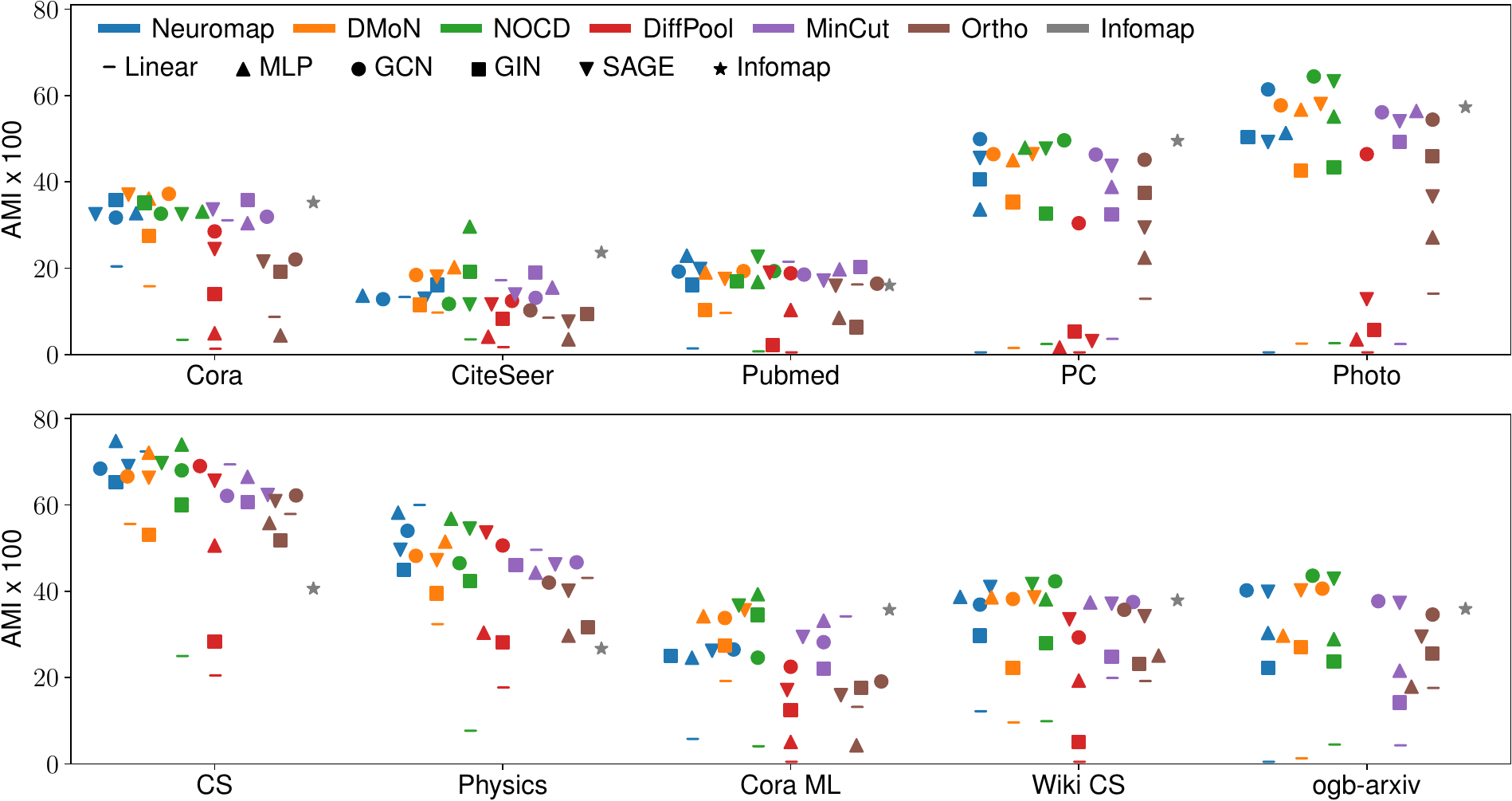}
	\caption{Average achieved AMI on real-world networks (higher is better) for $s = \left|Y\right|$, that is, the ``ground-truth'' number of communities. Colours indicate methods while shapes indicate neural network architectures. DiffPool ran out of memory on obg-arxiv.
	}
	\label{fig:real-world-results-ami-fixed-arch}
\end{figure*}

\subsection{Average Number of Detected Communities}
\begin{figure*}[h!]
	\centering
    \includegraphics[width=\linewidth]{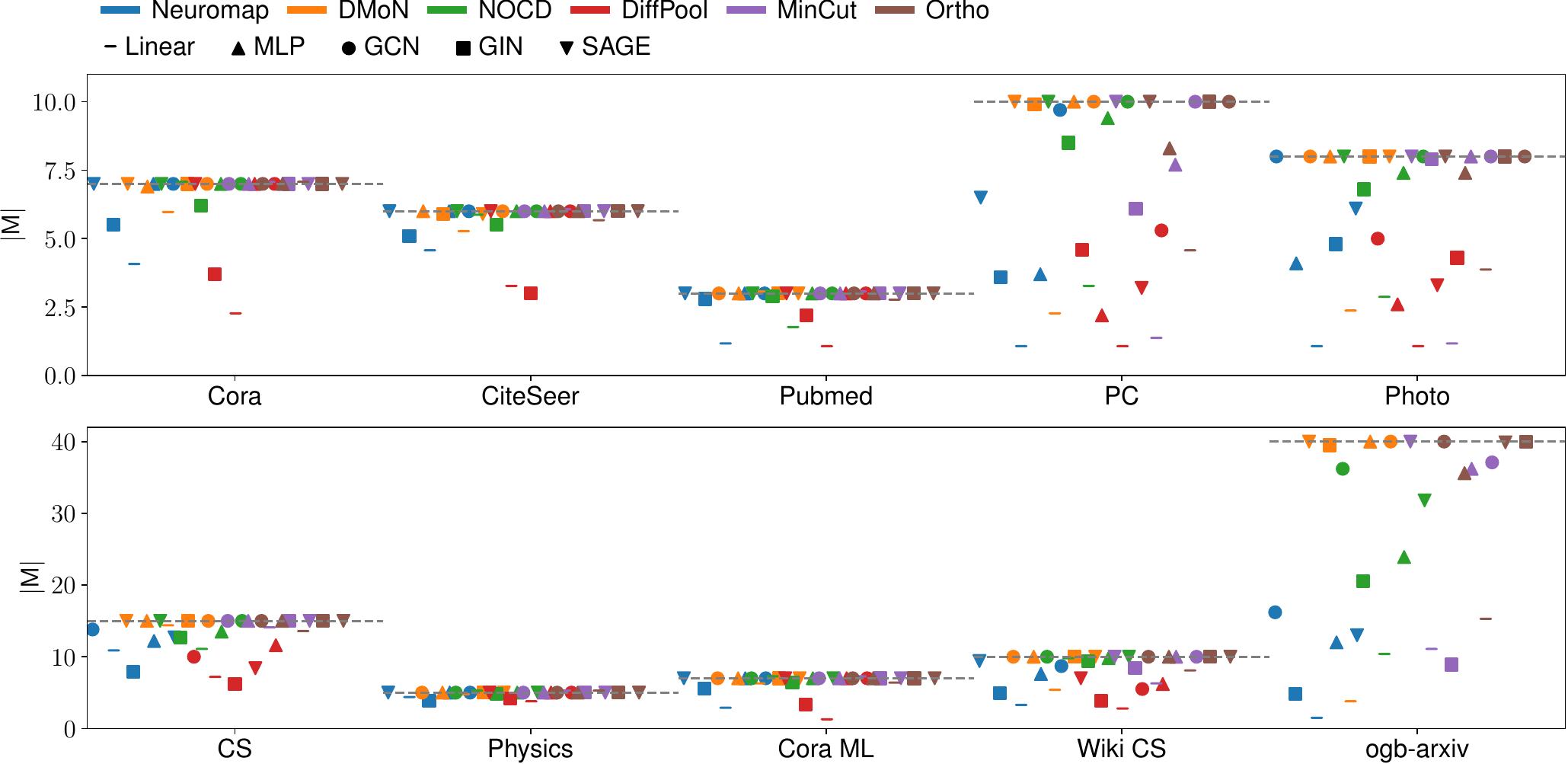}
	\caption{Average detected number of communities on real-world networks. Colours indicate methods while shapes indicate neural network architectures. The dashed horizontal lines show the correct number of communities for each dataset, which is also the maximum allowed number of communities. We omit Infomap in the plot due to the large number of detected communities.
	}
	\label{fig:real-world-results-m-fixed-arch}
\end{figure*}

\clearpage
\subsection{Tabulated AMI results on real-world networks}
Here, we report the detailed AMI performance for Neuromap, DMoN, NOCD, DiffPool, MinCut, Ortho, and Infomap on real-world networks for setting the number of hidden dimensions to $512$ and the maximum number of communities to $s = \left|Y\right|$.
When several of the best-performing methods achieve similar AMI, we use an independent two-sample t-test to determine whether one of them can be considered to perform better than the other.
In cases where their performances do not differ significantly, we mark both as best.
\begin{table*}[h!]
	\centering
	\caption{Average AMI in \% (higher is better) and their standard deviations on real-world networks.
	For each dataset, we highlight the best scores in bold and underline the second-best score.
    We tested each method with 2-layer MLP, GCN, GIN, and SAGE architectures, except for Infomap, which is not based on deep learning.
    OOM stands for ``out of memory''.
	}
    \resizebox{\linewidth}{!}{%
	\begin{tabular}{llcccccccccc}
        \toprule
		Method & Arch. & Cora & CiteSeer & PubMed & PC & Photo & CS & Physics & Cora ML & Wiki CS & ogb-arxiv \\
        \midrule
        Neuromap & LIN & $19.9 \pm 10.3$ & $12.8 \pm 6.9$ & $0.9 \pm 3.7$ & $0.0 \pm 0.0$ & $0.0 \pm 0.0$ & $71.9 \pm 3.7$ & $\tb{59.5 \pm 8.2}$ & $5.3 \pm 9.1$ & $11.7 \pm 11.0$ & $0.0 \pm 0.0$ \\
         & MLP & $32.7 \pm 4.6$ & $13.6 \pm 2.6$ & $\tb{22.9 \pm 4.6}$ & $33.6 \pm 13.3$ & $51.3 \pm 11.1$ & $\tb{74.8 \pm 2.4}$ & $\tb{58.2 \pm 6.0}$ & $24.6 \pm 2.9$ & $38.7 \pm 4.8$ & $30.3 \pm 1.8$ \\
         & GCN & $31.7 \pm 3.0$ & $12.8 \pm 2.9$ & $19.2 \pm 5.8$ & $\tb{49.9 \pm 3.6}$ & $61.4 \pm 4.2$ & $68.4 \pm 2.4$ & $54.0 \pm 6.4$ & $26.5 \pm 2.8$ & $36.9 \pm 4.2$ & $40.2 \pm 1.6$ \\
         & GIN & $\tb{35.8 \pm 4.1}$ & $16.1 \pm 3.2$ & $16.1 \pm 5.8$ & $40.5 \pm 3.9$ & $50.3 \pm 7.0$ & $65.2 \pm 3.4$ & $45.0 \pm 6.8$ & $25.0 \pm 3.5$ & $29.7 \pm 8.1$ & $22.2 \pm 2.5$ \\
         & SAGE & $32.5 \pm 5.0$ & $12.9 \pm 2.7$ & $19.8 \pm 4.6$ & $45.5 \pm 5.4$ & $49.2 \pm 6.1$ & $69.0 \pm 2.8$ & $49.6 \pm 6.0$ & $26.2 \pm 2.8$ & $\tb{41.0 \pm 2.5}$ & $39.9 \pm 1.6$ \\
        \midrule
        DMoN & LIN & $15.3 \pm 11.7$ & $9.2 \pm 6.9$ & $9.1 \pm 4.5$ & $1.0 \pm 0.8$ & $2.0 \pm 2.1$ & $55.1 \pm 12.9$ & $31.9 \pm 19.4$ & $18.7 \pm 12.0$ & $9.1 \pm 12.9$ & $0.8 \pm 1.1$ \\
         & MLP & $\tb{36.1 \pm 5.2}$ & $20.2 \pm 3.6$ & $19.0 \pm 6.5$ & $45.0 \pm 2.2$ & $56.7 \pm 3.1$ & $72.1 \pm 1.1$ & $51.5 \pm 3.9$ & $34.2 \pm 3.5$ & $38.6 \pm 3.0$ & $29.7 \pm 0.4$ \\
         & GCN & $\tb{37.2 \pm 4.2}$ & $18.4 \pm 2.6$ & $19.3 \pm 4.0$ & $46.4 \pm 1.9$ & $57.7 \pm 2.8$ & $66.6 \pm 1.9$ & $48.2 \pm 6.1$ & $33.8 \pm 3.5$ & $38.2 \pm 2.2$ & $40.6 \pm 0.4$ \\
         & GIN & $27.4 \pm 3.3$ & $11.5 \pm 3.1$ & $10.3 \pm 7.1$ & $35.3 \pm 2.3$ & $42.6 \pm 4.9$ & $53.1 \pm 3.7$ & $39.5 \pm 6.4$ & $27.4 \pm 2.4$ & $22.2 \pm 3.5$ & $27.1 \pm 1.6$ \\
         & SAGE & $\tb{37.0 \pm 4.0}$ & $18.0 \pm 3.4$ & $17.5 \pm 3.7$ & $46.4 \pm 1.6$ & $58.0 \pm 3.2$ & $66.3 \pm 1.6$ & $47.2 \pm 4.8$ & $\tu{35.6 \pm 3.8}$ & $38.6 \pm 2.1$ & $40.2 \pm 0.4$ \\
        \midrule
        NOCD & LIN & $2.9 \pm 1.3$ & $3.0 \pm 1.6$ & $0.2 \pm 0.5$ & $1.9 \pm 2.5$ & $2.1 \pm 3.0$ & $24.5 \pm 7.7$ & $7.2 \pm 4.5$ & $3.6 \pm 1.2$ & $9.4 \pm 2.5$ & $4.0 \pm 1.4$ \\
         & MLP & $\tb{33.1 \pm 14.7}$ & $\tb{29.6 \pm 10.8}$ & $16.8 \pm 8.1$ & $47.9 \pm 2.1$ & $55.1 \pm 15.4$ & $\tb{74.0 \pm 1.4}$ & $\tb{56.8 \pm 3.9}$ & $\tb{39.3 \pm 5.3}$ & $38.1 \pm 2.5$ & $28.9 \pm 0.5$ \\
         & GCN & $32.6 \pm 4.6$ & $11.7 \pm 1.7$ & $\tb{19.3 \pm 7.3}$ & $\tb{49.6 \pm 2.2}$ & $\tb{64.4 \pm 4.1}$ & $68.0 \pm 2.3$ & $46.5 \pm 7.3$ & $24.6 \pm 11.2$ & $\tb{42.3 \pm 2.2}$ & $\tb{43.6 \pm 0.5}$ \\
         & GIN & $\tb{35.1 \pm 4.8}$ & $19.2 \pm 3.4$ & $17.0 \pm 6.3$ & $32.6 \pm 3.7$ & $43.3 \pm 4.1$ & $60.0 \pm 3.8$ & $42.4 \pm 4.9$ & $34.5 \pm 3.9$ & $27.9 \pm 3.2$ & $23.7 \pm 1.2$ \\
         & SAGE & $32.5 \pm 3.7$ & $11.6 \pm 2.8$ & $\tb{22.6 \pm 4.9}$ & $47.7 \pm 2.4$ & $\tb{63.3 \pm 3.5}$ & $69.7 \pm 1.9$ & $54.5 \pm 4.7$ & $\tu{36.7 \pm 2.8}$ & $\tb{41.7 \pm 2.3}$ & $\tu{42.9 \pm 0.4}$ \\
        \midrule
        DiffPool & LIN & $0.8 \pm 1.9$ & $1.2 \pm 0.5$ & $0.0 \pm 0.0$ & $0.0 \pm 0.0$ & $0.0 \pm 0.0$ & $20.0 \pm 14.2$ & $17.2 \pm 18.2$ & $0.0 \pm 0.0$ & $0.0 \pm 0.0$ & OOM \\
         & MLP & $4.9 \pm 1.3$ & $4.1 \pm 1.4$ & $10.3 \pm 3.6$ & $1.6 \pm 0.4$ & $3.5 \pm 1.4$ & $50.6 \pm 6.7$ & $30.4 \pm 7.9$ & $5.1 \pm 1.0$ & $19.3 \pm 3.4$ & OOM \\
         & GCN & $28.5 \pm 4.1$ & $12.4 \pm 2.5$ & $18.8 \pm 5.5$ & $30.4 \pm 9.1$ & $46.4 \pm 6.5$ & $69.0 \pm 3.9$ & $50.6 \pm 6.7$ & $22.5 \pm 3.1$ & $29.3 \pm 4.8$ & OOM \\
         & GIN & $14.0 \pm 4.8$ & $8.2 \pm 3.1$ & $2.1 \pm 1.7$ & $5.3 \pm 5.6$ & $5.7 \pm 6.8$ & $28.4 \pm 9.1$ & $28.2 \pm 12.2$ & $12.5 \pm 3.8$ & $5.1 \pm 3.0$ & OOM \\
         & SAGE & $24.4 \pm 3.0$ & $11.6 \pm 2.5$ & $18.9 \pm 5.0$ & $3.1 \pm 0.6$ & $12.8 \pm 6.1$ & $65.6 \pm 4.5$ & $53.6 \pm 9.0$ & $17.1 \pm 2.9$ & $33.5 \pm 3.2$ & OOM \\
        \midrule
        MinCut & LIN & $30.6 \pm 4.3$ & $16.7 \pm 2.6$ & $\tb{21.0} \pm 1.9$ & $3.1 \pm 8.5$ & $1.9 \pm 6.4$ & $68.9 \pm 3.0$ & $49.1 \pm 3.4$ & $33.7 \pm 3.4$ & $19.4 \pm 10.4$ & $3.8 \pm 2.0$ \\
         & MLP & $30.4 \pm 4.5$ & $15.5 \pm 3.5$ & $\tb{19.7 \pm 6.3}$ & $38.8 \pm 13.3$ & $56.4 \pm 2.7$ & $66.5 \pm 1.2$ & $44.3 \pm 4.8$ & $33.2 \pm 4.7$ & $37.4 \pm 2.3$ & $21.6 \pm 8.1$ \\
         & GCN & $31.9 \pm 3.7$ & $13.1 \pm 2.8$ & $18.5 \pm 4.2$ & $46.3 \pm 2.0$ & $56.1 \pm 2.9$ & $62.1 \pm 1.4$ & $46.7 \pm 5.3$ & $28.2 \pm 3.2$ & $37.5 \pm 1.9$ & $37.7 \pm 1.4$ \\
         & GIN & $\tb{35.7 \pm 2.9}$ & $18.9 \pm 3.0$ & $20.2 \pm 4.1$ & $32.4 \pm 7.8$ & $49.2 \pm 4.9$ & $60.7 \pm 1.4$ & $46.1 \pm 4.3$ & $22.0 \pm 2.4$ & $24.8 \pm 11.9$ & $14.2 \pm 4.6$ \\
         & SAGE & $33.6 \pm 2.9$ & $13.9 \pm 3.5$ & $17.1 \pm 3.7$ & $43.7 \pm 2.5$ & $54.0 \pm 2.8$ & $62.3 \pm 1.2$ & $46.2 \pm 6.8$ & $29.4 \pm 3.7$ & $37.1 \pm 1.9$ & $37.3 \pm 0.5$ \\
        \midrule
        Ortho & LIN & $8.2 \pm 1.2$ & $8.0 \pm 2.3$ & $15.7 \pm 4.6$ & $12.4 \pm 6.5$ & $13.6 \pm 8.6$ & $57.4 \pm 2.6$ & $42.6 \pm 5.1$ & $12.7 \pm 3.7$ & $18.7 \pm 5.0$ & $17.1 \pm 1.1$ \\
         & MLP & $4.4 \pm 1.0$ & $3.5 \pm 1.0$ & $8.5 \pm 3.6$ & $22.4 \pm 5.7$ & $27.1 \pm 2.9$ & $55.8 \pm 2.7$ & $29.7 \pm 7.4$ & $4.3 \pm 1.1$ & $25.1 \pm 1.8$ & $17.9 \pm 0.5$ \\
         & GCN & $22.0 \pm 3.0$ & $10.2 \pm 2.1$ & $16.4 \pm 4.8$ & $45.1 \pm 1.9$ & $54.4 \pm 3.6$ & $62.2 \pm 2.0$ & $42.0 \pm 5.9$ & $19.1 \pm 3.2$ & $35.7 \pm 1.9$ & $34.6 \pm 0.8$ \\
         & GIN & $19.2 \pm 3.6$ & $9.4 \pm 2.9$ & $6.3 \pm 3.0$ & $37.4 \pm 3.8$ & $45.9 \pm 3.0$ & $51.8 \pm 2.7$ & $31.7 \pm 5.4$ & $17.6 \pm 2.6$ & $23.2 \pm 2.6$ & $25.5 \pm 1.2$ \\
         & SAGE & $21.5 \pm 4.1$ & $7.6 \pm 2.5$ & $15.9 \pm 3.8$ & $29.4 \pm 2.3$ & $36.6 \pm 3.2$ & $60.9 \pm 2.1$ & $40.1 \pm 6.3$ & $15.9 \pm 3.2$ & $34.2 \pm 2.2$ & $29.5 \pm 0.6$ \\
        \midrule
        Infomap &  & $35.2 \pm 0.2$ & $\tu{23.6 \pm 0.1}$ & $16.0 \pm 0.1$ & $\tb{49.5 \pm 0.5}$ & $57.3 \pm 1.1$ & $40.6 \pm 0.2$ & $26.7 \pm 0.1$ & $\tu{35.7 \pm 0.3}$ & $37.9 \pm 0.2$ & $35.9 \pm 0.1$ \\
        \bottomrule
	\end{tabular}
    }
	\label{table:real-ami-fixed-arch}
\end{table*}

\subsection{Significance of best Neuromap results vs. best baseline}
\begin{table}[h!]
	\centering
	\caption{Independent two-sample t-test between 25 samples of AMI values for Neuromap vs. the best baseline for 512 hidden features and $s = \left|Y\right|$. The p-values indicate for which datasets the null hypothesis ``the samples have the same mean'' can be rejected. \textcolor{palered}{Red p-values} highlight cases where the best baseline performs significantly better than Neuromap.}
	\begin{tabular}{llclcc}
        \toprule
		Dataset   & \multicolumn{2}{c}{Best Neuromap} & \multicolumn{2}{c}{Best Baseline} & p \\
		\midrule
		Cora & GIN & $35.8 \pm 4.1$ & DMoN GCN & $37.2 \pm 4.2$ & $0.25$ \\
		CiteSeer & GIN & $16.1 \pm 3.2$ & NOCD MLP & $29.6 \pm 10.8$ & $\fatred{2.4e^{-06}}$ \\
		PubMed & MLP & $22.9 \pm 4.6$ & NOCD SAGE & $22.6 \pm 4.9$ & $0.83$ \\
		PC & GCN & $49.9 \pm 3.6$ & NOCD GCN & $49.6 \pm 2.2$ & $0.73$ \\
		Photo & GCN & $61.4 \pm 4.2$ & NOCD GCN & $64.4 \pm 4.1$ & $0.02$ \\
		CS & MLP & $74.8 \pm 2.4$ & NOCD MLP & $74.0 \pm 1.4$ & $0.15$ \\
		Physics & LIN & $59.5 \pm 8.2$ & NOCD MLP & $56.8 \pm 3.9$ & $0.17$ \\
        Cora ML & GCN & $26.5 \pm 2.8$ & NOCD MLP & $39.3 \pm 5.3$ & $\fatred{1.6e^{-12}}$ \\
        Wiki CS & SAGE & $41.0 \pm 2.5$ & NOCD GCN & $42.3 \pm 2.2$ & $0.05$ \\
		ogb-arxiv & GCN & $40.2 \pm 1.6$ & NOCD GCN & $43.6 \pm 0.5$ & $\fatred{4.8e^{-11}}$ \\
        \bottomrule
	\end{tabular}
	\label{table:significance-best-neuromap-fixed-vs-baselines}
\end{table}

\clearpage
\subsection{Significance of Neuromap results vs. Infomap}
\begin{table*}[h!]
	\centering
	\caption{Independent two-sample t-test between 25 samples of AMI values for Neuromap with different (G)NN architectures vs. Infomap for $512$ hidden features and $s = \left|Y\right|$. The p-values indicate when the null hypothesis ``the samples have the same mean'' can be rejected. \textcolor{windowsblue}{Blue p-values} highlight cases where Neuromap performs significantly better than Infomap; \textcolor{palered}{red p-values} highlight cases where Infomap performs significantly better than Neuromap.
	}
    \begin{tabular}{lccccc}
        \toprule
		Dataset & LIN & MLP & GCN & GIN & SAGE \\
        \midrule
        Cora      & $\fatred{1.6e^{-7}}$ & $0.01$ & $\fatred{9.2e^{-6}}$ & $0.47$ & $0.02$ \\
        CiteSeer  & $\fatred{6.4e^{-8}}$ & $\fatred{1.0e^{-15}}$ & $\fatred{1.1e^{-15}}$ & $\fatred{2.7e^{-11}}$ & $\fatred{5.1e^{-16}}$ \\
        PubMed    & $\fatred{1.2e^{-16}}$ & $\fatblue{1.5e^{-7}}$ & $0.01$ & $0.95$ & $\fatblue{4.9e^{-4}}$ \\
        PC        & $\fatred{1.8e^{-48}}$ & $\fatred{5.1e^{-6}}$ & $0.55$ & $\fatred{4.4e^{-11}}$ & $\fatred{1.6e^{-3}}$ \\
        Photo     & $\fatred{1.8e^{-42}}$ & $0.02$ & $\fatblue{6.3e^{-5}}$ & $\fatred{5.7e^{-5}}$ & $\fatred{1.1e^{-6}}$ \\
        CS        & $\fatblue{6.2e^{-24}}$ & $\fatblue{1.8e^{-29}}$ & $\fatblue{3.0e^{-27}}$ & $\fatblue{2.5e^{-22}}$ & $\fatblue{8.8e^{-26}}$ \\
        Physics   & $\fatblue{3.1e^{-16}}$ & $\fatblue{6.5e^{-19}}$ & $\fatblue{7.3e^{-17}}$ & $\fatblue{1.6e^{-12}}$ & $\fatblue{7.2e^{-16}}$ \\
        Cora ML   & $\fatred{1.7e^{-14}}$ & $\fatred{5.3e^{-16}}$ & $\fatred{2.5e^{-14}}$ & $\fatred{9.9e^{-14}}$ & $\fatred{6.5e^{-15}}$ \\
        Wiki CS   & $\fatred{2.4e^{-11}}$ & $0.42$ & $0.29$ & $\fatred{5.3e^{-5}}$ & $\fatblue{3.4e^{-6}}$ \\
        ogb-arxiv & $\fatred{3.7e^{-62}}$ & $\fatred{4.5e^{-14}}$ & $\fatblue{2.0e^{-12}}$ & $\fatred{1.3e^{-19}}$ & $\fatblue{5.4e^{-12}}$ \\
        \bottomrule
	\end{tabular}
	\label{table:significance-neuromap-fixed-arch-vs-infomap}
\end{table*}

\subsection{Tabulated number of detected communities on real-world networks}
Here, we report the average number of detected communities for Neuromap, DMoN, NOCD, DiffPool, MinCut, Ortho, and Infomap on real-world networks for setting the number of hidden dimensions to $512$ and the maximum number of communities to $s = \left|Y\right|$.

\begin{table*}[h!]
	\centering
	\caption{Average number of detected communities and their standard deviations on real-world networks.
    We tested each method with 2-layer MLP, GCN, GIN, and SAGE architectures, except for Infomap, which is not based on deep learning.
    OOM stands for ``out of memory''.
	}
    \resizebox{\linewidth}{!}{%
	\begin{tabular}{llcccccccccc}
        \toprule
		Method & Arch. & Cora & CiteSeer & PubMed & PC & Photo & CS & Physics & Cora ML & Wiki CS & ogb-arxiv \\
        \midrule
        Neuromap & LIN & $4.0 \pm 1.2$ & $4.5 \pm 0.9$ & $1.1 \pm 0.3$ & $1.0 \pm 0.2$ & $1.0 \pm 0.0$ & $10.6 \pm 1.4$ & $4.1 \pm 0.7$ & $2.6 \pm 0.8$ & $3.0 \pm 0.9$ & $1.2 \pm 0.4$ \\
         & MLP & $7.0 \pm 0.0$ & $6.0 \pm 0.0$ & $3.0 \pm 0.0$ & $3.7 \pm 0.9$ & $4.1 \pm 1.1$ & $12.2 \pm 1.2$ & $5.0 \pm 0.0$ & $7.0 \pm 0.0$ & $7.6 \pm 1.1$ & $12.0 \pm 2.4$ \\
         & GCN & $7.0 \pm 0.0$ & $6.0 \pm 0.0$ & $3.0 \pm 0.0$ & $9.7 \pm 0.5$ & $8.0 \pm 0.2$ & $13.8 \pm 1.0$ & $5.0 \pm 0.0$ & $7.0 \pm 0.0$ & $8.7 \pm 0.9$ & $16.2 \pm 2.3$ \\
         & GIN & $5.5 \pm 0.8$ & $5.1 \pm 0.6$ & $2.8 \pm 0.4$ & $3.6 \pm 0.5$ & $4.8 \pm 0.9$ & $7.9 \pm 0.8$ & $3.9 \pm 0.7$ & $5.6 \pm 0.8$ & $4.9 \pm 1.2$ & $4.8 \pm 1.5$ \\
         & SAGE & $7.0 \pm 0.0$ & $6.0 \pm 0.0$ & $3.0 \pm 0.0$ & $6.5 \pm 0.9$ & $6.1 \pm 0.5$ & $12.7 \pm 1.2$ & $5.0 \pm 0.0$ & $7.0 \pm 0.0$ & $9.4 \pm 0.6$ & $13.0 \pm 1.1$ \\
        \midrule
        DMoN & LIN & $5.9 \pm 1.8$ & $5.2 \pm 1.5$ & $3.0 \pm 0.0$ & $2.2 \pm 0.9$ & $2.3 \pm 1.0$ & $14.1 \pm 2.5$ & $4.5 \pm 0.9$ & $6.0 \pm 1.8$ & $5.1 \pm 3.1$ & $3.5 \pm 3.2$ \\
         & MLP & $6.9 \pm 0.3$ & $6.0 \pm 0.0$ & $3.0 \pm 0.0$ & $10.0 \pm 0.0$ & $8.0 \pm 0.0$ & $15.0 \pm 0.0$ & $5.0 \pm 0.0$ & $7.0 \pm 0.2$ & $10.0 \pm 0.0$ & $40.0 \pm 0.0$ \\
         & GCN & $7.0 \pm 0.0$ & $6.0 \pm 0.0$ & $3.0 \pm 0.0$ & $10.0 \pm 0.0$ & $8.0 \pm 0.0$ & $15.0 \pm 0.0$ & $5.0 \pm 0.2$ & $7.0 \pm 0.2$ & $10.0 \pm 0.0$ & $40.0 \pm 0.0$ \\
         & GIN & $7.0 \pm 0.0$ & $5.9 \pm 0.3$ & $3.0 \pm 0.0$ & $9.9 \pm 0.3$ & $8.0 \pm 0.0$ & $15.0 \pm 0.0$ & $5.0 \pm 0.0$ & $7.0 \pm 0.0$ & $10.0 \pm 0.2$ & $39.5 \pm 1.0$ \\
         & SAGE & $7.0 \pm 0.2$ & $5.9 \pm 0.3$ & $3.0 \pm 0.0$ & $10.0 \pm 0.0$ & $8.0 \pm 0.0$ & $15.0 \pm 0.0$ & $5.0 \pm 0.0$ & $7.0 \pm 0.0$ & $10.0 \pm 0.0$ & $40.0 \pm 0.0$ \\
        \midrule
        NOCD & LIN & $7.0 \pm 0.0$ & $5.8 \pm 0.4$ & $1.7 \pm 0.5$ & $3.2 \pm 1.1$ & $2.8 \pm 1.2$ & $10.8 \pm 2.5$ & $5.0 \pm 0.0$ & $7.0 \pm 0.0$ & $9.5 \pm 0.7$ & $10.1 \pm 1.3$ \\
         & MLP & $7.0 \pm 0.0$ & $6.0 \pm 0.0$ & $3.0 \pm 0.0$ & $9.4 \pm 0.9$ & $7.4 \pm 0.6$ & $13.5 \pm 1.1$ & $5.0 \pm 0.0$ & $7.0 \pm 0.2$ & $9.8 \pm 0.6$ & $23.9 \pm 3.6$ \\
         & GCN & $7.0 \pm 0.0$ & $6.0 \pm 0.0$ & $3.0 \pm 0.0$ & $10.0 \pm 0.0$ & $8.0 \pm 0.0$ & $15.0 \pm 0.0$ & $5.0 \pm 0.0$ & $7.0 \pm 0.0$ & $10.0 \pm 0.0$ & $36.2 \pm 1.4$ \\
         & GIN & $6.2 \pm 0.6$ & $5.5 \pm 0.6$ & $2.9 \pm 0.3$ & $8.5 \pm 0.6$ & $6.8 \pm 1.0$ & $12.7 \pm 1.4$ & $4.8 \pm 0.4$ & $6.4 \pm 0.6$ & $9.4 \pm 0.6$ & $20.5 \pm 3.2$ \\
         & SAGE & $7.0 \pm 0.0$ & $6.0 \pm 0.0$ & $3.0 \pm 0.0$ & $10.0 \pm 0.2$ & $8.0 \pm 0.0$ & $15.0 \pm 0.0$ & $5.0 \pm 0.0$ & $7.0 \pm 0.0$ & $10.0 \pm 0.0$ & $31.8 \pm 1.8$ \\
        \midrule
        DiffPool & LIN & $2.2 \pm 0.7$ & $3.2 \pm 0.6$ & $1.0 \pm 0.0$ & $1.0 \pm 0.0$ & $1.0 \pm 0.0$ & $6.9 \pm 2.5$ & $3.5 \pm 1.0$ & $1.0 \pm 0.0$ & $2.5 \pm 0.6$ & OOM \\
         & MLP & $7.0 \pm 0.0$ & $6.0 \pm 0.0$ & $3.0 \pm 0.0$ & $2.2 \pm 0.4$ & $2.6 \pm 0.7$ & $11.6 \pm 1.9$ & $5.0 \pm 0.0$ & $7.0 \pm 0.0$ & $6.2 \pm 0.9$ & OOM \\
         & GCN & $7.0 \pm 0.0$ & $6.0 \pm 0.0$ & $3.0 \pm 0.0$ & $5.3 \pm 1.4$ & $5.0 \pm 0.7$ & $10.0 \pm 1.2$ & $5.0 \pm 0.0$ & $7.0 \pm 0.0$ & $5.5 \pm 1.0$ & OOM \\
         & GIN & $3.7 \pm 0.6$ & $3.0 \pm 0.7$ & $2.2 \pm 0.4$ & $4.6 \pm 1.2$ & $4.3 \pm 1.1$ & $6.2 \pm 3.1$ & $4.2 \pm 0.9$ & $3.3 \pm 0.9$ & $3.9 \pm 1.3$ & OOM \\
         & SAGE & $7.0 \pm 0.0$ & $6.0 \pm 0.0$ & $3.0 \pm 0.0$ & $3.2 \pm 0.4$ & $3.3 \pm 0.5$ & $8.4 \pm 1.3$ & $5.0 \pm 0.0$ & $7.0 \pm 0.0$ & $7.0 \pm 0.8$ & OOM \\
        \midrule
        MinCut & LIN & $7.0 \pm 0.0$ & $6.0 \pm 0.0$ & $3.0 \pm 0.0$ & $1.3 \pm 0.7$ & $1.1 \pm 0.3$ & $13.8 \pm 1.1$ & $5.0 \pm 0.0$ & $7.0 \pm 0.0$ & $6.0 \pm 2.4$ & $10.8 \pm 9.1$ \\
         & MLP & $7.0 \pm 0.0$ & $6.0 \pm 0.0$ & $3.0 \pm 0.0$ & $7.7 \pm 3.4$ & $8.0 \pm 0.0$ & $15.0 \pm 0.0$ & $5.0 \pm 0.0$ & $7.0 \pm 0.0$ & $10.0 \pm 0.0$ & $36.2 \pm 4.6$ \\
         & GCN & $7.0 \pm 0.0$ & $6.0 \pm 0.0$ & $3.0 \pm 0.0$ & $10.0 \pm 0.0$ & $8.0 \pm 0.0$ & $15.0 \pm 0.0$ & $5.0 \pm 0.0$ & $7.0 \pm 0.0$ & $10.0 \pm 0.0$ & $37.1 \pm 5.4$ \\
         & GIN & $7.0 \pm 0.0$ & $6.0 \pm 0.0$ & $3.0 \pm 0.0$ & $6.1 \pm 1.8$ & $7.9 \pm 0.3$ & $15.0 \pm 0.0$ & $5.0 \pm 0.0$ & $7.0 \pm 0.0$ & $8.4 \pm 3.4$ & $8.9 \pm 3.8$ \\
         & SAGE & $7.0 \pm 0.0$ & $6.0 \pm 0.0$ & $3.0 \pm 0.0$ & $10.0 \pm 0.0$ & $8.0 \pm 0.0$ & $15.0 \pm 0.0$ & $5.0 \pm 0.0$ & $7.0 \pm 0.0$ & $10.0 \pm 0.0$ & $40.0 \pm 0.0$ \\
        \midrule
        Ortho & LIN & $7.0 \pm 0.2$ & $5.6 \pm 0.5$ & $2.7 \pm 0.4$ & $4.5 \pm 1.5$ & $3.8 \pm 1.4$ & $13.3 \pm 0.6$ & $5.0 \pm 0.0$ & $6.1 \pm 1.4$ & $7.8 \pm 1.7$ & $15.0 \pm 1.7$ \\
         & MLP & $7.0 \pm 0.0$ & $6.0 \pm 0.0$ & $3.0 \pm 0.0$ & $8.3 \pm 1.5$ & $7.4 \pm 0.7$ & $15.0 \pm 0.0$ & $5.0 \pm 0.0$ & $7.0 \pm 0.0$ & $10.0 \pm 0.0$ & $35.6 \pm 1.8$ \\
         & GCN & $7.0 \pm 0.0$ & $6.0 \pm 0.0$ & $3.0 \pm 0.0$ & $10.0 \pm 0.0$ & $8.0 \pm 0.0$ & $15.0 \pm 0.0$ & $5.0 \pm 0.0$ & $7.0 \pm 0.0$ & $10.0 \pm 0.0$ & $40.0 \pm 0.0$ \\
         & GIN & $7.0 \pm 0.0$ & $6.0 \pm 0.0$ & $3.0 \pm 0.0$ & $10.0 \pm 0.0$ & $8.0 \pm 0.0$ & $15.0 \pm 0.0$ & $5.0 \pm 0.0$ & $7.0 \pm 0.0$ & $10.0 \pm 0.0$ & $40.0 \pm 0.0$ \\
         & SAGE & $7.0 \pm 0.0$ & $6.0 \pm 0.0$ & $3.0 \pm 0.0$ & $10.0 \pm 0.0$ & $8.0 \pm 0.0$ & $15.0 \pm 0.0$ & $5.0 \pm 0.0$ & $7.0 \pm 0.0$ & $10.0 \pm 0.0$ & $39.9 \pm 0.3$ \\
        \midrule
        Infomap &  & $282.2 \pm 4.1$ & $628.9 \pm 2.7$ & $924.9 \pm 9.1$ & $455.1 \pm 5.1$ & $223.5 \pm 4.4$ & $834.4 \pm 9.4$ & $1160.8 \pm 11.4$ & $287.4 \pm 3.8$ & $747.0 \pm 8.4$ & $4840.0 \pm 22.4$ \\
        \bottomrule
	\end{tabular}
    }
	\label{table:real-m-fixed-arch}
\end{table*}

\clearpage
\section{Performance impact of hidden layers and max. number of communities}\label{appx:performance-difference}
Here, we report the performance difference in terms of achieved average AMI score between using (i) $4\sqrt{n}$ hidden layers and a maximum of $s = \sqrt{n}$ communities, and (ii) $512$ hidden layers and a maximum of $s = \left|Y\right|$ communities, that is, the ``ground-truth'' number of communities.
The tabulated values show much better or worse the methods perform when using setup (ii) as compared to setup (i).
That is, a positive value, shown in green, means that setup (ii) gives a better result while a negative value, shown in red, means that setup (i) gives a better result.

We find that, for Neuromap, NOCD, and DiffPool, setup (i) generally works better.
In contrast, DMoN, MinCut, and Ortho often perform better with setup (ii), indicating that they require knowing the correct number of communities to perform well.
\begin{table*}[h!]
	\centering
	\caption{Performance difference in terms of achieved average AMI score between (i) $4\sqrt{n}$ hidden layers and a maximum of $s = \sqrt{n}$ communities, and (ii) $512$ hidden layers and a maximum of $s = \left|Y\right|$ communities, that is, the ``ground-truth'' number of communities.
    OOM stands for ``out of memory''.
	}
    \resizebox{\columnwidth}{!}{%
	\begin{tabular}{llcccccccccc}
        \toprule
		Method & Arch. & Cora & CiteSeer & PubMed & PC & Photo & CS & Physics & Cora ML & Wiki CS & ogb-arxiv \\
        \midrule
        Neuromap & LIN & $\textcolor{palered}{-12.00}$ & $\textcolor{palered}{-6.70}$ & $\textcolor{palered}{-16.30}$ & $0.00$ & $0.00$ & $\textcolor{palered}{-4.80}$ & $\textcolor{fadedgreen}{+3.00}$ & $\textcolor{palered}{-19.10}$ & $\textcolor{palered}{-30.20}$ & $\textcolor{palered}{-4.70}$ \\
         & MLP & $\textcolor{palered}{-3.90}$ & $\textcolor{palered}{-2.20}$ & $\textcolor{palered}{-0.30}$ & $\textcolor{fadedgreen}{+11.20}$ & $\textcolor{fadedgreen}{+11.50}$ & $\textcolor{palered}{-3.30}$ & $\textcolor{fadedgreen}{+2.30}$ & $\textcolor{palered}{-9.90}$ & $\textcolor{palered}{-5.20}$ & $\textcolor{palered}{-2.60}$ \\
         & GCN & $\textcolor{palered}{-6.80}$ & $\textcolor{palered}{-3.40}$ & $\textcolor{palered}{-1.60}$ & $\textcolor{palered}{-2.20}$ & $\textcolor{palered}{-3.60}$ & $\textcolor{palered}{-2.00}$ & $\textcolor{fadedgreen}{+3.70}$ & $\textcolor{palered}{-9.90}$ & $\textcolor{palered}{-4.20}$ & $\textcolor{palered}{-2.10}$ \\
         & GIN & $\textcolor{palered}{-10.30}$ & $\textcolor{palered}{-7.00}$ & $\textcolor{palered}{-8.50}$ & $\textcolor{palered}{-2.70}$ & $\textcolor{palered}{-7.90}$ & $\textcolor{palered}{-3.90}$ & $\textcolor{palered}{-7.70}$ & $\textcolor{palered}{-12.50}$ & $\textcolor{palered}{-5.20}$ & $\textcolor{palered}{-5.80}$ \\
         & SAGE & $\textcolor{palered}{-9.10}$ & $\textcolor{palered}{-5.40}$ & $\textcolor{palered}{-0.60}$ & $\textcolor{fadedgreen}{+1.90}$ & $\textcolor{fadedgreen}{+1.10}$ & $\textcolor{palered}{-1.90}$ & $\textcolor{palered}{-2.30}$ & $\textcolor{palered}{-12.90}$ & $\textcolor{palered}{-3.80}$ & $\textcolor{palered}{-3.70}$ \\
        \midrule
        DMoN & LIN & $\textcolor{fadedgreen}{+14.10}$ & $\textcolor{fadedgreen}{+6.00}$ & $\textcolor{fadedgreen}{+8.40}$ & $\textcolor{palered}{-2.40}$ & $\textcolor{palered}{-1.70}$ & $\textcolor{fadedgreen}{+29.90}$ & $\textcolor{fadedgreen}{+26.20}$ & $\textcolor{fadedgreen}{+18.30}$ & $\textcolor{fadedgreen}{+4.30}$ & $\textcolor{palered}{-3.20}$ \\
         & MLP & $\textcolor{palered}{-0.50}$ & $\textcolor{palered}{-2.30}$ & $\textcolor{fadedgreen}{+12.30}$ & $\textcolor{fadedgreen}{+29.60}$ & $\textcolor{fadedgreen}{+13.20}$ & $\textcolor{fadedgreen}{+31.90}$ & $\textcolor{fadedgreen}{+16.80}$ & $\textcolor{palered}{-2.00}$ & $\textcolor{fadedgreen}{+7.10}$ & $\textcolor{fadedgreen}{+16.90}$ \\
         & GCN & $\textcolor{palered}{-1.40}$ & $\textcolor{palered}{-2.00}$ & $\textcolor{fadedgreen}{+1.00}$ & $\textcolor{fadedgreen}{+2.60}$ & $\textcolor{fadedgreen}{+7.20}$ & $\textcolor{fadedgreen}{+13.50}$ & $\textcolor{fadedgreen}{+14.80}$ & $\textcolor{palered}{-3.90}$ & $\textcolor{fadedgreen}{+1.50}$ & $\textcolor{fadedgreen}{+2.20}$ \\
         & GIN & $\textcolor{palered}{-9.80}$ & $\textcolor{palered}{-8.00}$ & $\textcolor{palered}{-6.70}$ & $\textcolor{palered}{-0.10}$ & $\textcolor{palered}{-3.00}$ & $\textcolor{fadedgreen}{+3.20}$ & $\textcolor{fadedgreen}{+9.70}$ & $\textcolor{palered}{-7.30}$ & $\textcolor{palered}{-7.50}$ & $\textcolor{fadedgreen}{+0.90}$ \\
         & SAGE & $\textcolor{palered}{-2.10}$ & $\textcolor{palered}{-3.30}$ & $\textcolor{palered}{-0.20}$ & $\textcolor{fadedgreen}{+9.00}$ & $\textcolor{fadedgreen}{+7.20}$ & $\textcolor{fadedgreen}{+13.20}$ & $\textcolor{fadedgreen}{+13.60}$ & $\textcolor{palered}{-4.90}$ & $\textcolor{fadedgreen}{+2.20}$ & $\textcolor{fadedgreen}{+2.10}$ \\
        \midrule
        NOCD & LIN & $\textcolor{palered}{-2.20}$ & $\textcolor{fadedgreen}{+0.60}$ & $\textcolor{palered}{-3.50}$ & $\textcolor{fadedgreen}{+0.70}$ & $\textcolor{fadedgreen}{+2.00}$ & $\textcolor{palered}{-15.70}$ & $\textcolor{palered}{-12.10}$ & $\textcolor{fadedgreen}{+0.30}$ & $\textcolor{palered}{-1.40}$ & $\textcolor{palered}{-0.20}$ \\
         & MLP & $\textcolor{palered}{-13.00}$ & $\textcolor{palered}{-5.20}$ & $\textcolor{palered}{-3.70}$ & $\textcolor{fadedgreen}{+2.40}$ & $\textcolor{palered}{-5.40}$ & $\textcolor{fadedgreen}{+0.10}$ & $\textcolor{fadedgreen}{+6.60}$ & $\textcolor{palered}{-7.20}$ & $\textcolor{palered}{-2.80}$ & $\textcolor{fadedgreen}{+19.20}$ \\
         & GCN & $\textcolor{palered}{-8.50}$ & $\textcolor{palered}{-8.50}$ & $\textcolor{palered}{-1.90}$ & $\textcolor{palered}{-2.30}$ & $\textcolor{fadedgreen}{+1.20}$ & $\textcolor{fadedgreen}{+4.70}$ & $\textcolor{fadedgreen}{+4.50}$ & $\textcolor{palered}{-14.50}$ & $\textcolor{palered}{-1.90}$ & $\textcolor{palered}{-0.20}$ \\
         & GIN & $\textcolor{palered}{-8.30}$ & $\textcolor{palered}{-7.20}$ & $\textcolor{palered}{-4.30}$ & $\textcolor{palered}{-1.60}$ & $\textcolor{palered}{-12.00}$ & $\textcolor{palered}{-3.70}$ & $\textcolor{palered}{-5.90}$ & $\textcolor{palered}{-8.80}$ & $\textcolor{palered}{-9.70}$ & $\textcolor{palered}{-5.00}$ \\
         & SAGE & $\textcolor{palered}{-8.20}$ & $\textcolor{palered}{-8.20}$ & $\textcolor{fadedgreen}{+3.00}$ & $\textcolor{palered}{-3.40}$ & $\textcolor{palered}{-0.40}$ & $\textcolor{fadedgreen}{+6.00}$ & $\textcolor{fadedgreen}{+12.10}$ & $\textcolor{palered}{-3.90}$ & $\textcolor{palered}{-0.70}$ & $\textcolor{palered}{-0.20}$ \\
        \midrule
        DiffPool & LIN & $\textcolor{fadedgreen}{+0.40}$ & $\textcolor{fadedgreen}{+0.10}$ & $0.00$ & $0.00$ & $0.00$ & $\textcolor{palered}{-2.20}$ & $\textcolor{palered}{-1.70}$ & $0.00$ & $\textcolor{palered}{-2.40}$ & OOM \\
         & MLP & $\textcolor{palered}{-4.10}$ & $\textcolor{palered}{-2.60}$ & $\textcolor{palered}{-8.60}$ & $0.00$ & $\textcolor{fadedgreen}{+0.40}$ & $\textcolor{palered}{-13.90}$ & $\textcolor{palered}{-18.70}$ & $\textcolor{palered}{-4.40}$ & $\textcolor{palered}{-6.50}$ & OOM \\
         & GCN & $\textcolor{palered}{-11.30}$ & $\textcolor{palered}{-5.80}$ & $\textcolor{palered}{-4.90}$ & $\textcolor{palered}{-6.60}$ & $\textcolor{palered}{-7.40}$ & $\textcolor{palered}{-4.30}$ & $\textcolor{palered}{-11.80}$ & $\textcolor{palered}{-11.00}$ & $\textcolor{palered}{-6.40}$ & OOM \\
         & GIN & $\textcolor{palered}{-16.90}$ & $\textcolor{palered}{-13.10}$ & $\textcolor{palered}{-14.20}$ & $\textcolor{palered}{-9.20}$ & $\textcolor{palered}{-27.10}$ & $\textcolor{palered}{-21.90}$ & $\textcolor{palered}{-12.80}$ & $\textcolor{palered}{-19.10}$ & $\textcolor{palered}{-6.20}$ & OOM \\
         & SAGE & $\textcolor{palered}{-13.90}$ & $\textcolor{palered}{-4.80}$ & $\textcolor{palered}{-3.00}$ & $0.00$ & $\textcolor{palered}{-3.90}$ & $\textcolor{palered}{-5.80}$ & $\textcolor{palered}{-5.60}$ & $\textcolor{palered}{-12.50}$ & $\textcolor{palered}{-4.60}$ & OOM \\
        \midrule
        MinCut & LIN & $\textcolor{fadedgreen}{+4.50}$ & $\textcolor{fadedgreen}{+0.20}$ & $\textcolor{fadedgreen}{+18.80}$ & $\textcolor{fadedgreen}{+3.10}$ & $\textcolor{fadedgreen}{+1.90}$ & $\textcolor{fadedgreen}{+11.40}$ & $\textcolor{fadedgreen}{+12.30}$ & $\textcolor{fadedgreen}{+3.30}$ & $\textcolor{fadedgreen}{+3.40}$ & $0.00$ \\
         & MLP & $\textcolor{palered}{-4.30}$ & $\textcolor{palered}{-5.00}$ & $\textcolor{fadedgreen}{+14.60}$ & $\textcolor{fadedgreen}{+38.80}$ & $\textcolor{fadedgreen}{+51.20}$ & $\textcolor{fadedgreen}{+12.60}$ & $\textcolor{fadedgreen}{+14.90}$ & $\textcolor{palered}{-6.40}$ & $\textcolor{fadedgreen}{+9.30}$ & $\textcolor{fadedgreen}{+13.60}$ \\
         & GCN & $\textcolor{fadedgreen}{+5.70}$ & $\textcolor{fadedgreen}{+1.10}$ & $\textcolor{fadedgreen}{+7.50}$ & $\textcolor{fadedgreen}{+30.40}$ & $\textcolor{fadedgreen}{+10.90}$ & $\textcolor{fadedgreen}{+17.40}$ & $\textcolor{fadedgreen}{+14.50}$ & $\textcolor{fadedgreen}{+2.00}$ & $\textcolor{fadedgreen}{+5.40}$ & $\textcolor{fadedgreen}{+3.80}$ \\
         & GIN & $\textcolor{palered}{-3.60}$ & $0.00$ & $\textcolor{fadedgreen}{+0.40}$ & $\textcolor{fadedgreen}{+25.60}$ & $\textcolor{fadedgreen}{+13.70}$ & $\textcolor{fadedgreen}{+28.00}$ & $\textcolor{palered}{-5.60}$ & $\textcolor{fadedgreen}{+15.10}$ & $\textcolor{fadedgreen}{+23.50}$ & $\textcolor{fadedgreen}{+3.90}$ \\
         & SAGE & $\textcolor{fadedgreen}{+2.10}$ & $\textcolor{palered}{-2.30}$ & $\textcolor{fadedgreen}{+1.60}$ & $\textcolor{fadedgreen}{+2.10}$ & $\textcolor{fadedgreen}{+7.10}$ & $\textcolor{fadedgreen}{+15.20}$ & $\textcolor{fadedgreen}{+15.70}$ & $\textcolor{palered}{-6.50}$ & $\textcolor{fadedgreen}{+4.80}$ & $\textcolor{fadedgreen}{+4.30}$ \\
        \midrule
        Ortho & LIN & $\textcolor{fadedgreen}{+1.20}$ & $\textcolor{fadedgreen}{+2.40}$ & $\textcolor{palered}{-0.80}$ & $\textcolor{palered}{-6.40}$ & $\textcolor{palered}{-11.10}$ & $\textcolor{fadedgreen}{+12.00}$ & $\textcolor{fadedgreen}{+15.90}$ & $\textcolor{palered}{-2.90}$ & $\textcolor{palered}{-10.90}$ & $\textcolor{palered}{-2.00}$ \\
         & MLP & $\textcolor{palered}{-0.90}$ & $\textcolor{palered}{-0.30}$ & $\textcolor{palered}{-2.10}$ & $\textcolor{fadedgreen}{+2.60}$ & $\textcolor{palered}{-3.50}$ & $\textcolor{palered}{-0.50}$ & $\textcolor{palered}{-2.30}$ & $\textcolor{palered}{-1.00}$ & $\textcolor{palered}{-3.70}$ & $\textcolor{palered}{-0.60}$ \\
         & GCN & $\textcolor{palered}{-0.90}$ & $\textcolor{palered}{-0.70}$ & $\textcolor{fadedgreen}{+2.60}$ & $\textcolor{fadedgreen}{+4.40}$ & $\textcolor{fadedgreen}{+8.40}$ & $\textcolor{fadedgreen}{+10.60}$ & $\textcolor{fadedgreen}{+9.60}$ & $\textcolor{palered}{-0.10}$ & $\textcolor{fadedgreen}{+3.60}$ & $\textcolor{fadedgreen}{+0.30}$ \\
         & GIN & $\textcolor{palered}{-11.50}$ & $\textcolor{palered}{-8.40}$ & $\textcolor{palered}{-7.30}$ & $\textcolor{fadedgreen}{+3.40}$ & $\textcolor{fadedgreen}{+3.60}$ & $\textcolor{fadedgreen}{+7.70}$ & $\textcolor{fadedgreen}{+6.70}$ & $\textcolor{palered}{-11.10}$ & $\textcolor{palered}{-6.20}$ & $\textcolor{fadedgreen}{+1.40}$ \\
         & SAGE & $\textcolor{fadedgreen}{+3.50}$ & $\textcolor{palered}{-0.80}$ & $\textcolor{fadedgreen}{+6.00}$ & $\textcolor{palered}{-6.00}$ & $\textcolor{palered}{-9.70}$ & $\textcolor{fadedgreen}{+8.10}$ & $\textcolor{fadedgreen}{+5.70}$ & $\textcolor{palered}{-0.40}$ & $\textcolor{fadedgreen}{+2.50}$ & $\textcolor{fadedgreen}{+4.60}$ \\
        \midrule
        Infomap &  & $0.00$ & $0.00$ & $0.00$ & $0.00$ & $0.00$ & $0.00$ & $0.00$ & $0.00$ & $0.00$ & $0.00$ \\
        \bottomrule
	\end{tabular}
    }
	\label{table:performance-difference}
\end{table*}

\clearpage
\section{Further results on a synthetic network with overlapping communities\label{appx:more-overlapping}}
Here we show the detected communities on a synthetic network with overlapping communities with a dense linear layer, a 2-layer MLP, a 2-layer GIN, and a 2-layer SAGE architecture. We use $\left|V\right|$ hidden channels and set the maximum number of communities to $s \in \left\{2,3\right\}$.

\subsection{Linear-layer-based results}
\begin{figure*}[h!]
	\centering
    \vspace*{1\baselineskip}
	\begin{overpic}[width=.99\linewidth]{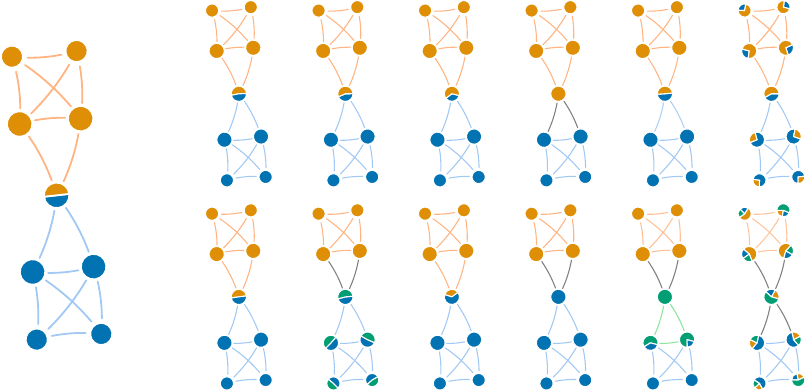}
		\put(3,45){True}
        \put(18,36){$s=2$}
        \put(18,11){$s=3$}
        \put(23.5,51){Neuromap}
        \put(38.5,51){DMoN}
        \put(51.5,51){NOCD}
        \put(64,51){DiffPool}
        \put(78,51){MinCut}
        \put(92,51){Ortho}
	\end{overpic}
	\caption{
		Linear-layer-based results on a synthetic network with overlapping communities where the leftmost network shows the true community structure.
        Nodes are drawn as pie charts to visualise their community assignments.
        The top and bottom rows show results for a maximum of $s = 2$ and $s = 3$ communities, respectively.
	}
	\label{fig:alcides_lin}
\end{figure*}

\subsection{MLP-based results}
\begin{figure*}[h!]
	\centering
    \vspace*{1\baselineskip}
	\begin{overpic}[width=.99\linewidth]{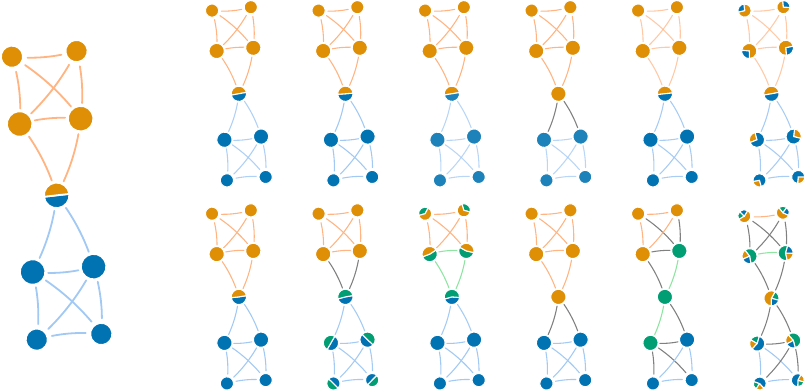}
		\put(3,45){True}
        \put(18,36){$s=2$}
        \put(18,11){$s=3$}
        \put(23.5,51){Neuromap}
        \put(38.5,51){DMoN}
        \put(51.5,51){NOCD}
        \put(64,51){DiffPool}
        \put(78,51){MinCut}
        \put(92,51){Ortho}
	\end{overpic}
	\caption{
		MLP-based results on a synthetic network with overlapping communities where the leftmost network shows the true community structure.
        Nodes are drawn as pie charts to visualise their community assignments.
        The top and bottom rows show results for a maximum of $s = 2$ and $s = 3$ communities, respectively.
	}
	\label{fig:alcides_mlp}
\end{figure*}

\clearpage
\subsection{GIN-based results}
\begin{figure*}[h!]
	\centering
    \vspace*{1.5\baselineskip}
	\begin{overpic}[width=\linewidth]{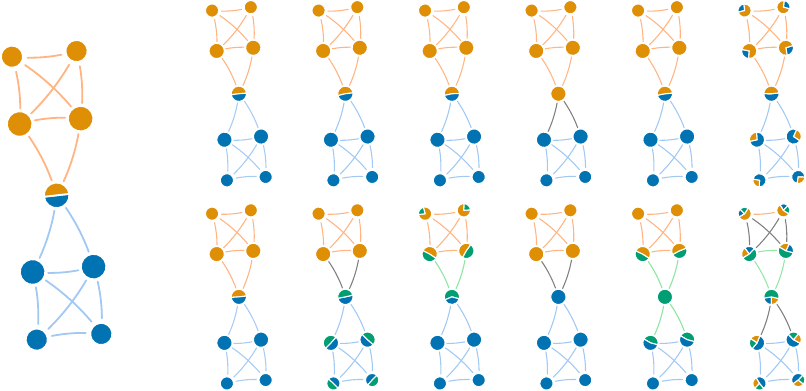}
		\put(3,45){True}
        \put(18,36){$s=2$}
        \put(18,11){$s=3$}
        \put(23.5,51){Neuromap}
        \put(38.5,51){DMoN}
        \put(51.5,51){NOCD}
        \put(64,51){DiffPool}
        \put(78,51){MinCut}
        \put(92,51){Ortho}
	\end{overpic}
	\caption{
		GIN-based results on a synthetic network with overlapping communities where the leftmost network shows the true community structure.
        Nodes are drawn as pie charts to visualise their community assignments.
        The top and bottom rows show results for a maximum of $s = 2$ and $s = 3$ communities, respectively.
	}
	\label{fig:alcides_gin}
\end{figure*}

\subsection{SAGE-based results}
\begin{figure*}[h!]
	\centering
    \vspace*{1.5\baselineskip}
	\begin{overpic}[width=\linewidth]{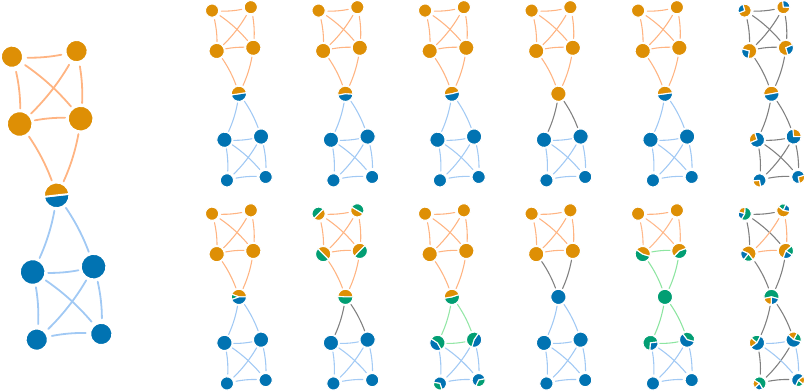}
		\put(3,45){True}
        \put(18,36){$s=2$}
        \put(18,11){$s=3$}
        \put(23.5,51){Neuromap}
        \put(38.5,51){DMoN}
        \put(51.5,51){NOCD}
        \put(64,51){DiffPool}
        \put(78,51){MinCut}
        \put(92,51){Ortho}
	\end{overpic}
	\caption{
		SAGE-based results on a synthetic network with overlapping communities where the leftmost network shows the true community structure.
        Nodes are drawn as pie charts to visualise their community assignments.
        The top and bottom rows show results for a maximum of $s = 2$ and $s = 3$ communities, respectively.
	}
	\label{fig:alcides_sage}
\end{figure*}

\end{document}